\documentclass{article}
\usepackage[T1]{fontenc}         % tectonic runs XeTeX: without T1 the style's times falls back to Latin Modern
\usepackage{iclr2027_conference,times}
\usepackage{natbib}
\usepackage{microtype}
\usepackage{amsmath,amssymb,amsthm}
\usepackage{booktabs}
\usepackage{array}               % ragged-right p-columns in the arm-name table
\usepackage{graphicx}
\usepackage{hyperref}
\usepackage{url}
\usepackage{xspace}
\usepackage{algorithm}
\usepackage{algorithmic}
\usepackage{tikz}                % method figure, sections_v5/fig_method.tex
\usepackage{placeins}            % appendix: flush each subsection's floats before the next heading
\usetikzlibrary{arrows.meta,positioning,calc}

\newtheorem{observation}{Observation}
\newtheorem{proposition}{Proposition}

\newcommand{\arm}[1]{\textsc{#1}\xspace}
\newcommand{\Ascratch}{\arm{from scratch}}                                   % random
\newcommand{\Aproj}{\arm{dense projection}}                                  % projection
\newcommand{\Asel}{\arm{selection}}                                          % subclone
\newcommand{\Aiso}{\arm{selection~+ read-out rescale}}                       % subclone_iso
\newcommand{\Ars}{\arm{selection~+ rescale}}                                 % subclone_rs
\newcommand{\Acomp}{\arm{selection~+ compensation}}                          % hybrid
\newcommand{\Acomprs}{\arm{selection~+ compensation~+ rescale}}              % hybrid_rs
\newcommand{\Ainfrs}{\arm{informed selection~+ rescale}}                     % subclone_rs_act
\newcommand{\Ainfcomprs}{\arm{informed selection~+ compensation~+ rescale}}  % hybrid_rs_act
\newcommand{\Ssel}{\arm{sel.}}
\newcommand{\Siso}{\arm{sel.+read-out resc.}}
\newcommand{\Srs}{\arm{sel.+resc.}}
\newcommand{\Scomp}{\arm{sel.+comp.}}
\newcommand{\Scomprs}{\arm{sel.+comp.+resc.}}
\newcommand{\Sinfrs}{\arm{inf.\ sel.+resc.}}
\newcommand{\Sinfcomprs}{\arm{inf.\ sel.+comp.+resc.}}

\newif\ifappendixfloats\appendixfloatsfalse
\newcommand{\AlgCompSel}{%
\begin{algorithm}[!htb]
\footnotesize
\caption{Compensated Selection: \Acomprs (code name \textsc{hybrid\_rs};
App.~Table~\ref{tab:armnames})}
\label{alg:compsel}
\begin{algorithmic}[1]
\REQUIRE donor weights $W_L$; frozen-corpus activation variances and second
moments $\Sigma$; target shape (residual $d'$, head dim, MLP width, block map)
\ENSURE installable small-model weights (no target-model access)
\STATE $\mathcal{R} \gets$ top-$d'$ residual features by mean-over-layer activation variance \COMMENT{one global basis}
\FOR{each kept block $b$}
  \STATE $\mathcal{K}_{qk} \gets$ frequency-matched rotary pairs (stride $2$) $\cup$ top non-rotary dims by $\lVert q\rVert{+}\lVert k\rVert$
  \STATE $\mathcal{K}_{vo} \gets$ top head dims by $\lVert V_{\text{row}}\rVert\cdot\lVert O_{\text{col}}\rVert$ \COMMENT{whole heads if head count shrinks}
  \STATE $\mathcal{K}_{mlp} \gets$ top units by $\lVert\text{up}_{\text{row}}\rVert\cdot\lVert\text{down}_{\text{col}}\rVert$
  \STATE slice $W_L$ block $b$ to $(\mathcal{R},\mathcal{K}_{qk},\mathcal{K}_{vo},\mathcal{K}_{mlp})$; LayerNorms and biases ride along
  \STATE $O' \gets \text{LScomp}(O,\,\Sigma_{\text{attn-in}},\,\mathcal{K}_{vo})$;\; $\text{MLP\_down}' \gets \text{LScomp}(\text{MLP\_down},\,\Sigma_{\text{mlp-in}},\,\mathcal{K}_{mlp})$ \COMMENT{Eq.~\ref{eq:lscomp}, function lever}
  \IF{rescale}
    \STATE multiply LN-fronted read-in paths $Q,K,V,\text{MLP\_up}$ (and output embedding) by $\sqrt{d_{\text{in}}/d_{\text{in}}'}$ \COMMENT{dynamics lever}
  \ENDIF
\ENDFOR
\RETURN assembled small model
\end{algorithmic}
\end{algorithm}
}
\newcommand{\TabPredictor}{%
\begin{table}[!htb]
\centering
\footnotesize
\caption{Tests 3--4 (learnability and transfer). Patch-predictor error reduction
over predicting zero vs.\ a label-shuffled control; 1000-sample bootstrap CI on
the real$-$control gap. No predictor beats zero or separates from its control.}
\label{tab:predictor}
\begin{tabular}{llcccc}
\toprule
Setting & Predictor & Reduction & Shuffle & Gap CI$_{95}$ & $p$ \\
\midrule
Held-out layers (1.4B$\to$410M) & linear  & $-0.0004$ & $-0.0004$ & $[-0.0000,-0.0000]$ & 0.99 \\
                                & MLP-512 & $-0.0003$ & $-0.0003$ & $[-0.0000,-0.0000]$ & 1.00 \\
\midrule
Transfer (410M$\to$160M)        & linear  & $-0.0004$ & $-0.0004$ & --- & 1.00 \\
                                & MLP-512 & $-0.0003$ & $-0.0002$ & --- & 1.00 \\
\bottomrule
\end{tabular}
\\[3pt]
{\footnotesize CIs are reported for the in-pair held-out test only; the transfer
test reports $p$ (M4). Reductions are negative because no predictor improves on
predicting zero.}
\end{table}
}

\newsavebox{\tikzfitbox}

\renewenvironment{tikzpicture}[1][]
  {\begin{lrbox}{\tikzfitbox}\begin{origtikzpicture}[#1]}
  {\end{origtikzpicture}\end{lrbox}%
   \ifdim\wd\tikzfitbox>\textwidth
     \resizebox{\textwidth}{!}{\usebox{\tikzfitbox}}%
   \else\usebox{\tikzfitbox}\fi}

\iclrfinalcopy  % arXiv preprint: de-anonymized (real author), no line numbers.

\title{Wiring Beats Blending: Structure-Aware Compensation\\for Transformer Downscaling}

\author{Ravi Satya Durga Prasad Yenugula\\
\texttt{Independent Researcher}}

\begin{document}
\maketitle

\begin{abstract}
Model families are trained size by size. Can a pretrained large model instead be
converted into a smaller sibling? We characterize the 1.4B$\to$410M conversion
in Pythia end to end. Representations align strongly across sizes (ridge
$R^2=0.84$); parameters align weakly. Dense weight projection is destructive; a
bit-exact control places the fault in basis mixing, which breaks rotary, per-head, GELU, and LayerNorm structure. Residuals after the best-fit linear
operator carry no learnable or transferable signal under shuffle controls, so
conversion value lives in initialization. Matched-budget continued pre-training
separates two independent levers, least-squares compensation (function lever,
best zero-shot) and variance-preserving rescale (dynamics lever, best
endpoints), with a placement rule read off the architecture: compensation is
well-posed exactly where no normalization sits between cut and read; norm-fronted paths take rescale. At 30M tokens compensation beats the best subcloning
variant on a width-reduced pair ($84.6\pm2.1$ vs. $90.7\pm3.8$, 6/6 seeds) and a
held-out depth-reduced pair (109.3 vs. 117.9, 3/3 seeds); the two reach parity
at $33\times$ the budget, and at ~$5\times$ the donor scale the levers anti-synergize. A second family, Qwen3-1.7B$\to$0.6B (RMSNorm, SwiGLU, grouped-query
attention), takes the placement rule unchanged but reverses a $2\times2$
crossing the re-fit with the activation statistic it reads: on Pythia the re-fit
carries the gain and the statistic adds 2.3 perplexity; on Qwen3, with far more
heterogeneous MLP unit energies, the statistic carries 66 points and the re-fit
is a null on top. Informed selection + compensation + rescale is the best arm in
both families, $19\times$ and $23\times$ lower perplexity than from scratch at
matched budget (2.94 and 3.13 nats), and beats structured pruning plus
distillation. Code, checkpoints, and the frozen
evaluation corpus are released at
\url{https://github.com/rsdpyenugula/ScaleOp} and
\url{https://huggingface.co/YenugulaAIML/scaleop-pythia-conversions}.
\end{abstract}

\section{Introduction}
\label{sec:intro}

Transformer families ship as ladders of discrete sizes (Pythia alone spans
70M to 12B parameters~\citep{biderman2023pythia}), and every rung is bought
separately: each size is a full pretraining run, even though the members are
trained on identical data with a shared tokenizer and architecture and differ
only in width and depth. Recent work derives a smaller model from a larger one
by structured pruning~\citep{xia2024sheared,ma2023llmpruner} and knowledge
distillation~\citep{hinton2015distilling}, often in one
pipeline~\citep{muralidharan2024minitron}, but keeps the donor in the loop: the
pruning mask is scored against it and the teacher is queried throughout
training. We study the cheaper channel in which the donor is used once, at
initialization, and then discarded, and compare against the prune-plus-distill
recipe at matched budget (\S\ref{sec:race:distill}). What actually transfers
between two sizes of the same pretrained family, and through what operation
can it be carried into the smaller shape?

We answer end to end on the Pythia suite, where identical data and tokenizer
isolate size as the only variable (\S\ref{sec:setup}): locate the cross-size
relation (\S\ref{sec:representations}), diagnose why dense projection of the
large weights fails (\S\ref{sec:projection}), test whether what the best-fit
linear operator misses is learnable (\S\ref{sec:residuals}), and, since it is
not, run a matched-budget recovery race on Pythia and on a second family,
Qwen3, leading with the result and the recommendation
(\S\ref{sec:race:result}, Figure~\ref{fig:method}) and following with the
evidence, from the ladder that decomposes conversion into two independent
levers to the hardening across seeds, budgets, pairs, family
(\S\ref{sec:race:family}), and scale (\S\ref{sec:race}). Our contributions:
\begin{itemize}
\item \textbf{A controlled characterization of what transfers across sizes.}
  Within a family trained on identical data, representations align strongly
  under a linear map (held-out ridge $R^2=0.84$) while parameters align weakly
  ($R^2=0.25$--$0.39$ in the cleanest, embedding case)
  (\S\ref{sec:representations}).
\item \textbf{A mechanically verified diagnosis of why dense projection fails.}
  Rebuilding the donor from its own extracted weights reproduces its logits
  bit-exactly, so the projected model's collapse is attributable to basis mixing
  alone: dense maps break rotary pairing, per-head attention, elementwise GELU,
  and per-feature LayerNorm, the last the largest single factor in the
  zero-shot collapse (\S\ref{sec:projection}).
\item \textbf{A null result under controls.} After the best-fit linear operator,
  weight residuals carry no cross-layer direction, no patch-learnable signal, and
  no cross-pair transfer (four tests vs.\ shuffle controls with bootstrap CIs,
  $p\ge0.99$), so no learned zero-shot correction exists at this granularity
  (\S\ref{sec:residuals}).
\item \textbf{A two-lever decomposition of conversion, and a method that stacks
  them.} The levers partition by where normalization sits: least-squares
  compensation (the classical post-pruning re-fit, \S\ref{sec:related}) is
  well-posed only on the paths with no norm between the cut and the read; the
  norm-fronted paths take the reference recipe's variance-preserving scalar
  rescale.
  Compensation is the function lever (best zero-shot), rescale the
  training-dynamics lever (best endpoints), and their combination, Compensated
  Selection, dominates the strongest variant of weight
  subcloning~\citep{samragh2023weight} on both axes in every seed
  ($84.6\pm2.1$ vs.\ $90.7\pm3.8$ final perplexity, 6/6 paired wins; zero-shot
  $18.5$k vs.\ $61.9$k), ends at $18\times$ lower perplexity than from scratch
  at 30M tokens ($2.91$ nats), holds out-of-domain and at $2\times$ the
  training context, and stays $\sim$13$\times$ below from scratch ($2.59$
  nats) on a held-out depth-dominated pair (\S\ref{sec:race}).
\item \textbf{A second family, and a family-dependent split.} On
  Qwen3-1.7B$\to$0.6B (RMSNorm, SwiGLU, grouped-query attention) the placement
  rule applies unchanged and the best arm ends at $23\times$ lower perplexity
  than from scratch ($3.13$ nats), but a $2\times2$ that crosses the re-fit
  with the activation
  statistic it reads comes out reversed: on Pythia the re-fit carries the gain
  and the statistic adds $2.3$ perplexity; on Qwen3 the statistic carries $66$
  points and the re-fit is a null on top of it ($+0.3\pm0.6$). A per-block
  diagnostic and a proposition tie the split to how unequal the MLP unit
  energies are; the arm that stacks informed selection, compensation, and
  rescale is the best in both families (\S\ref{sec:race:family}).
\end{itemize}

\section{Related work}
\label{sec:related}

\paragraph{Downscaling pretrained models.}
The closest prior work is Weight Subcloning~\citep{samragh2023weight}, which
initializes a smaller transformer by ranking units of a larger pretrained one,
copying selected rows and columns, and rescaling. Sheared
LLaMA~\citep{xia2024sheared} learns pruning masks jointly with continued
pre-training; we instead hold the training recipe fixed and isolate what the
initialization alone contributes. Knowledge distillation~\citep{hinton2015distilling},
often combined with pruning~\citep{muralidharan2024minitron}, transfers function
through a teacher's outputs during a full training run; we study what transfers
through weights at negligible cost, and compare against and combine with
distillation in \S\ref{sec:race:distill}. Table~\ref{tab:positioning} places
these methods on the four axes this paper varies.

\begin{table}[t]
\centering
\scriptsize
\caption{Positioning on the four axes this paper varies; cells come only from
the characterizations in this section, ``---'' where the cited description
does not fix an axis.}
\label{tab:positioning}
\setlength{\tabcolsep}{2pt}
\renewcommand{\arraystretch}{1.05}
\begin{tabular}{@{}p{0.56in}p{0.62in}p{0.6in}p{0.63in}p{0.62in}@{}}
\toprule
Method & Selection statistic & Repair at the cut & Repair placement & Post-cut training \\
\midrule
OBS \citeyearpar{hassibi1993optimal} & second-order saliency & closed-form weight update & remaining weights & --- \\
LLM-Pruner \citeyearpar{ma2023llmpruner} & donor-scored importance & --- & --- & --- \\
Sheared LLaMA \citeyearpar{xia2024sheared} & learned masks & --- & --- & continued pre-training, joint with the masks \\
Minitron \citeyearpar{muralidharan2024minitron} & donor-scored importance & --- & --- & distillation from the donor throughout \\
Weight subcloning \citeyearpar{samragh2023weight} & activation magnitude & scalar $\sqrt{d/d'}$ rescale & every matrix & continued pre-training \\
GRAIL \citeyearpar{tang2026grail} & any (selector-agnostic) & ridge re-fit (Gram matrix) & FFN down; attention out & none (post-hoc) \\
SliceGPT \citeyearpar{ashkboos2024slicegpt} & PCA of the residual stream & PCA rotation, exact under RMSNorm & residual axis only & none \\
This work & weight norm, or activation energy & ridge re-fit (Eq.~\ref{eq:lscomp}) $+$ $\sqrt{d/d'}$ & re-fit where no norm fronts the read; rescale elsewhere & continued pre-training; donor discarded \\
\bottomrule
\end{tabular}
\end{table}

\paragraph{Growing pretrained models.}
The mirror direction has a longer history: Net2Net~\citep{chen2016net2net}
introduced function-preserving widening and deepening,
bert2BERT~\citep{chen2022bert2bert} adapted it to transformers, staged
training~\citep{shen2022staged} formalized loss-preserving growth operators,
and LiGO~\citep{wang2023ligo} learns a structured linear map from small to
large weights.

\paragraph{Representation similarity and parameter symmetries.}
We use linear CKA~\citep{kornblith2019similarity} descriptively and held-out
ridge $R^2$ as the decisive alignment measure (\S\ref{sec:representations}).
Our selection-versus-blending result is an
instance of the principle behind permutation-symmetry work on model
merging~\citep{ainsworth2023git} and
stitching~\citep{lenc2015understanding,bansal2021revisiting}: networks compare
naturally modulo the transformations their architecture is invariant to, and
structured selection (a permutation respecting head boundaries and rotary
frequency pairs~\citep{su2024roformer}, then coordinate deletion) stays inside
that group where dense projection leaves it.

\paragraph{Positioning.}
The re-fit is shared prior art. Least-squares reconstruction after a cut runs
from Optimal Brain Surgeon~\citep{hassibi1993optimal} and channel
pruning~\citep{he2017channel} through LLM-Pruner~\citep{ma2023llmpruner} to,
most directly, GRAIL~\citep{tang2026grail}: a post-hoc, zero-finetuning,
selector-agnostic step that forms the Gram matrix of the consumer's input on a
small calibration set, fits a ridge map from the kept hidden units to the
original ones, and absorbs it into the consumer weights at the FFN
down-projection and the attention output projection, the construction of
Eq.~\ref{eq:lscomp} at the same two points. Our contribution is not the
estimator but (i)~the family-conversion setting, where residual width and
depth are cut too, so most reads sit behind a normalization, which GRAIL,
pruning at fixed residual width with no training after the cut, does not
face; (ii)~the placement rule for where the re-fit is well-posed and what to
do elsewhere (rescale); (iii)~the two-lever decomposition and its stacking;
(iv)~the information-vs-re-fit control and its family dependence; (v)~the
scale failure map and the representation/parameter alignment diagnosis.
SliceGPT~\citep{ashkboos2024slicegpt} rotates the residual stream by PCA,
exploiting the computational invariance of RMSNorm networks, then deletes rows
and columns with no fine-tuning: a dense basis change
kept exactly where it is invariant, consistent with our diagnosis that mixing
the structured axes (rotary pairs, heads, elementwise units) is what breaks
(\S\ref{sec:projection}). Our subcloning baseline re-implements
\citet{samragh2023weight} with three disclosed differences (hence
\emph{subcloning-style}): (a)~heads and MLP units are scored by weight norm
and residual lanes by activation variance, where they use activation
magnitudes (the \emph{informed} arms of \S\ref{sec:race:family} restore an
activation statistic); (b)~when depth shrinks we drop blocks at even stride
rather than from the middle; (c)~their $\sqrt{d/d'}$ rescale is evaluated as
a separate lever (\S\ref{sec:race:levers}) rather than applied to every
matrix, where it collapses zero-shot quality.

\section{Experimental setup}
\label{sec:setup}

\paragraph{Model families and conversion pairs.}
We study the Pythia suite~\citep{biderman2023pythia}: GPT-NeoX models trained on
identical data (the Pile) with a shared tokenizer, isolating size as the
only variable between family members. Our primary pair converts
\textbf{1.4B}$\,\to\,$\textbf{410M} (24 layers and 16 heads in both; only widths
shrink: residual $2048\!\to\!1024$, head dim $128\!\to\!64$, MLP
$8192\!\to\!4096$), so layer correspondence is one-to-one and the analysis
isolates width. A held-out pair, \textbf{410M}$\,\to\,$\textbf{160M}, exercises
the other axes: depth halves ($24\!\to\!12$ blocks), the head count drops
($16\!\to\!12$; head dim unchanged), and width cuts are mild ($\sim$25\%). A
third pair, \textbf{6.9B}$\,\to\,$\textbf{1.4B}, tests scale (all three axes
reduced; \S\ref{sec:race:scale}). A fourth pair comes from a second family,
\textbf{Qwen3-1.7B-Base}$\,\to\,$\textbf{Qwen3-0.6B-Base}~\citep{yang2025qwen3}
(RMSNorm~\citep{zhang2019rmsnorm}, gated SwiGLU~\citep{shazeer2020glu},
grouped-query attention~\citep{ainslie2023gqa} with QK-norm, tied embeddings;
28 layers and the tokenizer shared; hidden $2048\!\to\!1024$, MLP intermediate
$6144\!\to\!3072$), a width-only cut like the primary pair
(\S\ref{sec:race:family}).

\paragraph{Frozen evaluation corpus.}
All representation analyses probe every model with the same inputs: 10{,}000
sequences of 128 tokens from the Pile (\texttt{pile-uncopyrighted}), tokenized
once and frozen for the project (for Qwen3, the same stream head under its own
tokenizer); the first 500 sequences (all token positions) form the similarity
subset (\S\ref{sec:representations}), the rest contribute mean-pooled and
last-token activations. Continued pre-training streams the same corpus but
skips the frozen-set region (contamination guard), and each seed trains on a
disjoint stream offset.

\paragraph{Behavioral evaluation.}
Language-modeling quality is strided sliding-window perplexity on WikiText-103
validation (every token scored once); hardening evaluations add C4 perplexity
(out-of-domain), perplexity at $2\times$ the training context (2048), and
zero-shot accuracy on LAMBADA, ARC-Easy, HellaSwag, and PIQA via
\texttt{lm-eval}. Reference models anchor every table and match published
Pythia numbers. All forward passes are greedy and seeded, bit-identical on
repetition, and weight extraction is validated by exact reconstruction (a model
rebuilt from its own extracted weights reproduces the original logits with zero
difference, in both families; \S\ref{sec:projection}), so downstream quality
changes are attributable to conversion mathematics, not assembly.

\paragraph{Training protocol.}
Every conversion arm within an experiment shares identical data order,
optimizer, cosine schedule, and token budget, so arms differ only in their
starting weights, and every init is read off the donor and the frozen corpus
alone; hyperparameters, budgets, and hardware are in Appendix~\ref{app:protocol}.

\section{Representations align, parameters do not}
\label{sec:representations}

On the primary width-only pair (1.4B$\,\to\,$410M) we look for the cross-size
relation in the activations the two models compute and in the parameters they
store, and find it in only one (App.~Table~\ref{tab:align}; method in
Appendix~\ref{app:align}).

\paragraph{Activations are largely a linear image across widths; raw parameters are not.}
Per layer we fit a large$\,\to\,$small map on pooled activations (8k train /
2k held-out) and score held-out $R^2$. A full linear map (ridge,
$2048\!\to\!1024$) reaches a mean held-out $R^2=0.844$: the 410M pooled
representation is largely a linear projection of the 1.4B's. Procrustes
(rotation plus one global scale) already reaches $0.717$, so most of the
relation is a change of basis and the $\sim$0.13 gap is non-orthogonal
reshaping; both maps dip together at layers 3--5, the layers linear CKA also
flags as distinct (biased estimator of \citet{kornblith2019similarity}, 500
sequences $\times$ all positions: diagonal mean $0.883$ but a saturated middle
band, so the claim rests on the held-out $R^2$, not on CKA;
App.~Fig.~\ref{fig:align}). The token
embeddings are the one weight matrix where the two sizes share an axis (the
vocabulary), so a single map applies (rows split $80/20$); even in this
cleanest case a linear map explains only $0.25$--$0.39$ of weight variance
(ridge $R^2=0.248$ input, $0.386$ output, against $0.844$ for activations). Projection (\S\ref{sec:projection}) must
therefore leave a substantial residual; whether it is learnable or noise
(\S\ref{sec:residuals}) decides where conversion value comes from.

\section{Why dense projection fails}
\label{sec:projection}

We build a projection of the large weights into the small shape from the size
relationship alone (never the target's weights), measure its residual, run
the assembled model, and diagnose why it breaks.

\paragraph{Target-free projection.}
Each weight is mapped by a pair of structured linear operators,
\begin{equation}
\hat{W} = P_{\text{out}}\, W\, P_{\text{in}}^{\top},
\quad
P_{\bullet} \in \{\, P_{\text{res}},\, P_{QK},\, P_{VO},\, P_{\text{MLP}} \,\},
\label{eq:project}
\end{equation}
with a global activation-PCA basis $P_{\text{res}}$ on the residual axis and
SVD bases shared between interacting weights on the internal axes ($Q,K$ share
one, $V,O$ one, \texttt{MLP\_UP}/\texttt{MLP\_DOWN} a joint basis from
$[\mathrm{UP},\,\mathrm{DOWN}^{\top}]$), so composed operations stay
consistent.

\paragraph{The projected weights miss the target solution.}
Against the actual 410M weights (relative Frobenius error, mean over 24
layers; App.~Table~\ref{tab:proj}) every target-free error exceeds $1$, worse
than predicting zero: the projected weights share almost nothing with the
410M's particular solution. Even a shared operator $(A,B)$ fitted with access
to the target, a diagnostic upper bound, only reaches $0.66$--$0.80$, so at
best $\sim$30\% of weight variance is linearly explainable across sizes with
one shared map.

\paragraph{Behavioral anchors.}
A 410M-shaped model assembled from the projected 1.4B weights does not beat
random init zero-shot (strided WikiText-103 perplexity,
App.~Table~\ref{tab:anchor}): random init is near-uniform (perplexity on the
order of vocabulary size), the projected model confidently wrong, far worse
than uniform; and since the unprojected rebuild reproduces the 1.4B's logits
with $\max|\text{diff}| = 0$ (\S\ref{sec:setup}), the entire loss is
attributable to Eq.~\ref{eq:project}.

\paragraph{Structure mixing breaks the computation.}
A dense basis change on any internal axis silently changes what the block
computes, because every structured or nonlinear operation is tied to specific
coordinates: rotary embeddings act on fixed per-head dimension pairs, so
head-space mixing distorts positions; attention is per-head, yet a global
$2048\!\to\!1024$ head-space basis mixes dimensions across heads; GELU is
elementwise, so $\mathrm{GELU}(Bx)\neq B\,\mathrm{GELU}(x)$; and LayerNorm gains
and biases are per-feature, so a rotated stream matches no diagonal gain, which
is why the projected-LN variant explodes to $10^{13}$.

\paragraph{Separating the violations.}
Replacing the projected model's mismatched norms with neutral ones (gain $1$,
bias $0$) cuts zero-shot perplexity $\sim$$15\times$ ($181{,}242 \to 12{,}323$,
$2.69$ nats; App.~Table~\ref{tab:anchor}, Table~\ref{tab:ladder}; strided and
quick evaluations respectively, which agree at these magnitudes), so LayerNorm
mismatch dominates the immediate collapse. What remains is the basis-mixing
cost, and it is a trainability cost more than a zero-shot one: with neutral
norms the projected model starts within $7\%$ of structure-preserving
selection ($12{,}323$ vs.\ $11{,}509$ at $t{=}0$) yet ends at $3\times$ the
perplexity after identical training ($1{,}054.5$ vs.\ $355.8$ at 30M tokens,
$1.09$ nats; Table~\ref{tab:ladder}; rotary, per-head, and GELU mixing are not
individually isolated). Zero-shot conversion therefore needs
structure-respecting selection (whole coordinates, so rotary pairs, heads,
units, and per-feature norms stay valid); dense projection serves only as an
initialization for brief fine-tuning (\S\ref{sec:race}).

\section{Post-operator residuals carry no learnable signal}
\label{sec:residuals}

Even the strongest shared linear operator explains at best $\sim$30\% of weight
variance (relative error $0.66$--$0.80$; \S\ref{sec:projection}). Is the
residual a structured, learnable correction, or noise? We define it per weight
as
\begin{equation}
\Delta(l,\text{type}) \;=\; W_{410\text{M}} \;-\; A\, W_{1.4\text{B}}\, B^{\top},
\label{eq:delta}
\end{equation}
with the shared fitted operator $(A,B)$ per type (the strongest-alignment
condition of \S\ref{sec:projection}), so whatever remains no single linear
size-map explains, and test the null that $\Delta$ is exchangeable noise with
the observed marginal statistics: four tests compare a statistic of $\Delta$
with the same statistic under a matched control (a per-type shuffle, or a
shape- and scale-matched Gaussian), with a 1000-sample bootstrap confidence
interval on the real$-$control gap for the decisive tests.

\paragraph{Tests 1--2: spectrum and cross-layer consistency.}
The effective rank of $\Delta$ sits $2.8$--$6.4\%$ below both controls in
every type (App.~Table~\ref{tab:erank}, Fig.~\ref{fig:spectra}): a faint
spectral concentration. The
mean pairwise cosine of vectorized $\Delta$ across layers is $\approx\pm0.0002$
in every type, identical to the shuffle control: no direction is shared across
layers, so no single correction is uniformly missing.

\paragraph{Tests 3--4: learnability (decisive) and transfer.}
A patch predictor $R(\hat{W}\text{ patch} + \text{type/layer/position})
\to \Delta\text{ patch}$ is trained on 18 layers and scored on 6 stratified
held-out layers as error reduction over predicting zero; the control retrains
on $\Delta$ shuffled across layers. Neither a linear nor an MLP-512 predictor
beats zero (reductions $-0.0004$/$-0.0003$), and real and control are
indistinguishable (bootstrap gap CI$_{95}\approx[-0.0000,-0.0000]$, $p\ge0.99$;
App.~Table~\ref{tab:predictor}). Trained on all primary-pair layers and
evaluated on the held-out 410M$\,\to\,$160M pair, the same predictors reduce
error by $-0.0004$/$-0.0003$, again indistinguishable from the shuffle control
($p=1.00$): nothing transfers either.

\paragraph{Handoff to \S\ref{sec:race}.}
The faint spectral deficit is a generic statistical trace, not a
predictor-exploitable correspondence (cross-check in
Appendix~\ref{app:reconcile}); on every test that could exploit structure we
fail to reject the null, so conversion value must come from initialization
plus brief fine-tuning, the matched-budget race of \S\ref{sec:race}.

\ifappendixfloats\else\TabPredictor\fi  % deferred to appendix in the page-limited build

\section{Matched-budget conversion: two independent levers}
\label{sec:race}

% Method figure (\S\ref{sec:race:result}): (a) the pre-norm block drawn once,
% with its cut axes and where each lever goes; (b) the two-family 2x2. Every
% number is one of Table~\ref{tab:qwen3main}'s. Input from 7_race.tex; as a
% figure* it lands at the top of the page after the one it is called from.
\definecolor{rescC}{RGB}{20,110,60}
\definecolor{compC}{RGB}{170,40,30}
\definecolor{cutC}{RGB}{40,70,150}
\definecolor{pyC}{RGB}{25,80,160}
\definecolor{qwC}{RGB}{200,90,15}
\begin{figure*}[t]
\centering
\begin{tikzpicture}[
  font=\scriptsize,
  >={Latex[length=1.5mm,width=1.1mm]},
  box/.style={draw, rounded corners=1pt, minimum height=4mm, inner sep=2pt, align=center, fill=white},
  norm/.style={box, fill=black!8},
  sum/.style={draw, circle, inner sep=0pt, minimum size=2.8mm, fill=white},
  tag/.style={rounded corners=1pt, inner sep=1.2pt, font=\tiny, align=center},
  resc/.style={tag, draw=rescC, fill=rescC!12, text=rescC},
  comp/.style={tag, draw=compC, fill=compC!12, text=compC},
  cut/.style={font=\tiny, text=cutC},
  fam/.style={font=\tiny, anchor=north west, text width=9.35cm, align=left, inner sep=0pt},
  flow/.style={->, semithick},
  cell/.style={draw, rounded corners=1.5pt, minimum width=2.1cm, minimum height=0.85cm, align=left, inner sep=2pt, fill=white},
  eff/.style={font=\tiny, inner sep=0.6pt},
]
% ---------------- (a) one pre-norm block ----------------
\node[font=\scriptsize\bfseries, anchor=north west] at (-0.15,3.55) {(a)};
\coordinate (rin) at (0,0);
\coordinate (tap) at (0.55,0);
\node[sum] (s1) at (5.35,0) {$+$};
\node[sum] (s2) at (6.05,0) {$+$};
\node[norm] (nf) at (7.3,0) {final norm};
\node[box] (lm) at (8.7,0) {lm\_head};
\draw[line width=1.4pt] (rin) -- (s1);
\draw[line width=1.4pt] (s1) -- (s2);
\draw[line width=1.4pt, ->] (s2) -- (nf);
\draw[flow] (nf) -- (lm);
\fill (tap) circle (1.1pt);
\draw[line width=1.4pt] (tap) -- (0.55,2.8);
\node[cut, anchor=north] at (2.95,-0.1) {residual stream, width $d\to d'$ (cut in both families)};
% attention branch
\node[norm] (n1) at (1.25,1.25) {norm};
\node[box] (qkv) at (2.3,1.25) {$Q\;K\;V$};
\node[box] (att) at (3.5,1.25) {attention};
\node[box] (o) at (4.55,1.25) {$O$};
\draw[flow] (0.55,1.25) -- (n1);
\draw[flow] (n1) -- (qkv);
\draw[flow] (qkv) -- (att);
\draw[flow] (att) -- (o);
\draw[flow] (o.east) -| (s1);
\node[cut, anchor=north] at (2.9,1.02) {heads $\times$ head dim};
% MLP branch
\node[norm] (n2) at (1.25,2.8) {norm};
\node[box] (up) at (2.35,2.8) {up\,/\,gate};
\node[box] (act) at (3.45,2.8) {act};
\node[box] (dn) at (4.45,2.8) {down};
\draw[flow] (0.55,2.8) -- (n2);
\draw[flow] (n2) -- (up);
\draw[flow] (up) -- (act);
\draw[flow] (act) -- (dn);
\draw[flow] (dn.east) -| (s2);
\node[cut, anchor=north] at (2.9,2.57) {MLP hidden units $h\to h'$};
% lever placement
\node[resc, above=1pt of qkv] {rescale};
\node[resc, above=1pt of up] {rescale};
\node[resc, above=1pt of lm] {rescale};
\node[comp, above=1pt of o] {compensate\\(if head dims cut)};
\node[comp, above=1pt of dn] {compensate};
% the two families
\node[fam] at (0,-0.58) {\textcolor{pyC}{\textbf{Pythia}} (LayerNorm, GELU,
  parallel residual): residual width, head dims, and MLP units are cut, so
  $O$ and down are compensated; $Q$, $K$, $V$, up, and the output embedding
  are rescaled.};
\node[fam] at (0,-1.11) {\textcolor{qwC}{\textbf{Qwen3}} (RMSNorm, SwiGLU,
  grouped-query attention, sequential residual): heads stay intact, so only
  down is compensated; $Q$, $K$, $V$, gate, up, and the tied
  \texttt{lm\_head} (via the final norm's gain) are rescaled.};
% ---------------- (b) the two-family 2x2 ----------------
\begin{scope}[shift={(10.1,-0.15)}]
\node[font=\scriptsize\bfseries, anchor=north west] at (-0.15,3.7) {(b)};
\node[font=\tiny, anchor=south] at (4.0,3.28) {selection statistic};
\node[anchor=south] at (2.4,2.98) {weight norm};
\node[anchor=south] at (5.6,2.98) {activation energy};
\node[font=\tiny, rotate=90, anchor=south] at (0.1,1.7) {re-fit};
\node[anchor=east] at (1.2,2.5) {none};
\node[anchor=east, align=right] at (1.2,0.85) {compen-\\sation};
\node[cell] (c11) at (2.4,2.5) {\textcolor{pyC}{Pythia $89.7\pm3.7$}\\\textcolor{qwC}{Qwen3 $119.6\pm11.8$}};
\node[cell] (c12) at (5.6,2.5) {\textcolor{pyC}{Pythia $87.4\pm2.2$}\\\textcolor{qwC}{Qwen3 $53.4\pm0.9$}};
\node[cell] (c21) at (2.4,0.85) {\textcolor{pyC}{Pythia $84.0\pm1.8$}\\\textcolor{qwC}{Qwen3 $88.9\pm7.0$}};
\node[cell, line width=0.9pt] (c22) at (5.6,0.85) {\textcolor{pyC}{Pythia $\mathbf{80.2\pm2.3}$}\\\textcolor{qwC}{Qwen3 $\mathbf{53.1\pm1.3}$}};
% statistic effect (across the rows); arrow width = drop / family's largest drop
\draw[->, pyC, line width=0.8pt]  ([yshift=3pt]c11.east)  -- ([yshift=3pt]c12.west)  node[eff, midway, above, text=pyC] {$-2.3$};
\draw[->, qwC, line width=1.9pt]  ([yshift=-3pt]c11.east) -- ([yshift=-3pt]c12.west) node[eff, midway, below, text=qwC] {$-66.2$};
\draw[->, pyC, line width=1.1pt]  ([yshift=3pt]c21.east)  -- ([yshift=3pt]c22.west)  node[eff, midway, above, text=pyC] {$-3.7$};
\draw[->, qwC, line width=1.15pt] ([yshift=-3pt]c21.east) -- ([yshift=-3pt]c22.west) node[eff, midway, below, text=qwC] {$-35.9$};
% re-fit effect (down the columns)
\draw[->, pyC, line width=1.55pt] ([xshift=-3pt]c11.south) -- ([xshift=-3pt]c21.north) node[eff, midway, left, text=pyC] {$-5.7$};
\draw[->, qwC, line width=1.05pt] ([xshift=3pt]c11.south)  -- ([xshift=3pt]c21.north)  node[eff, midway, right, text=qwC] {$-30.7$};
\draw[->, pyC, line width=1.9pt]  ([xshift=-3pt]c12.south) -- ([xshift=-3pt]c22.north) node[eff, midway, left, text=pyC] {$-7.2$};
\draw[->, qwC, line width=0.3pt]  ([xshift=3pt]c12.south)  -- ([xshift=3pt]c22.north)  node[eff, midway, right, text=qwC] {$-0.3$};
\end{scope}
\end{tikzpicture}
\caption{\textbf{Where each lever goes, and which one carries the gain.}
\textbf{(a)}~One pre-norm block, its three cut axes, and the read that
consumes each cut input. A read behind a normalization ($Q$, $K$, $V$, up,
gate, the output embedding through the final norm) has no fixed least-squares
target and takes the rescale; a norm-free read of a cut input ($O$ when head
dimensions are cut, MLP down always) is re-fit by compensation
(Eq.~\ref{eq:lscomp}). \textbf{(b)}~The $2\times2$ of re-fit (rows) by
selection statistic (columns): final perplexity of both families per cell and
the paired drop on each arrow (width relative to the family's largest). On
Pythia the large drops run down the columns (re-fit), on Qwen3 across the rows
(statistic), yet the same corner, \Ainfcomprs, is best in both. All cells are
30M-token WikiText-103 finals, mean$\pm$std over three data-draw seeds
(Table~\ref{tab:qwen3main}).}
\label{fig:method}
\end{figure*}

If conversion value lives in initialization, the decisive test is a
matched-budget recovery race: same target shape, data, schedule, and token
budget for every candidate, and continued pre-training reveals which
starting weights end lowest. The init decomposes into two independent
levers, least-squares \textbf{compensation}, which fixes what the init computes
(best zero-shot), and variance-preserving \textbf{rescale}, which fixes the
scale the optimizer sees (best endpoints).

\subsection{Result and recommendation}
\label{sec:race:result}

\begin{table}[t]
\centering
\footnotesize
\caption{Two families, one protocol: final WikiText-103 perplexity at 30M
tokens, mean$\pm$std over three data-draw seeds (Pythia \Ascratch: single
seed), crossing the selection statistic (norm, or energy: \emph{informed})
with the re-fit; paired effects on the arrows of Figure~\ref{fig:method}b and
in App.~Tables~\ref{tab:seeded}, \ref{tab:qwen3seeded}, \ref{tab:paired}.
Short forms and code names in App.~Table~\ref{tab:armnames}; best per column
in \textbf{bold}.}
\label{tab:qwen3main}
\scriptsize
\setlength{\tabcolsep}{1.8pt}
\begin{tabular}{llrr}
\toprule
Arm & stat.\,/\,re-fit & Pythia 1.4B$\to$410M & Qwen3 1.7B$\to$0.6B \\
\midrule
\Ascratch   & --- / ---     & 1{,}519.0    & $1{,}217.7\pm30.7$ \\
\Srs        & norm / no     & $89.7\pm3.7$ & $119.6\pm11.8$ \\
\Scomprs    & norm / yes    & $84.0\pm1.8$ & $88.9\pm7.0$ \\
\Sinfrs     & energy / no   & $87.4\pm2.2$ & $53.4\pm0.9$ \\
\Sinfcomprs & energy / yes  & $\mathbf{80.2\pm2.3}$ & $\mathbf{53.1\pm1.3}$ \\
\bottomrule
\end{tabular}
\end{table}

Figure~\ref{fig:method} and Table~\ref{tab:qwen3main} carry the result:
\emph{informed} selection scores the V/O and MLP units it keeps by their
activation energy on the frozen corpus instead of by weight norm, and stacked with
compensation and rescale it is the best arm in both families, \Ainfcomprs:
$80.2\pm2.3$ on Pythia 1.4B$\to$410M and $53.1\pm1.3$ on Qwen3-1.7B$\to$0.6B,
$19\times$ and $23\times$ lower perplexity than \Ascratch at matched budget
($2.94$ and $3.13$ nats). Where each lever goes is read off the wiring
(Figure~\ref{fig:method}a): compensation where a norm-free read consumes a cut
input, rescale on every read a normalization fronts. Which lever carries the
gain is a family property (Figure~\ref{fig:method}b): on Pythia the re-fit
($5.7$ and $7.2$ perplexity, $3/3$ seeds each; the statistic adds $2.3$), on
Qwen3 the statistic ($66.2$ points; the re-fit on top of it is a null,
$+0.3\pm0.6$). The recommendation is the same in both: select by activation
energy, compensate the norm-free reads, rescale the rest.

\subsection{The ablation ladder and its two levers}
\label{sec:race:ladder}

Table~\ref{tab:ladder} walks a seven-arm ladder from an untrained model to the
full Pythia method: monotone in the endpoint (each rung improves the final
perplexity) but not in zero-shot, and the two levers explain that dissonance.
Bare \Asel (355.8) ends at $4.3\times$ lower perplexity than \Ascratch
(1{,}519.0) at matched compute ($1.45$ nats), and \Aproj (1{,}054.5), despite
its scrambled wiring, still ends $1.4\times$ below from scratch ($0.36$
nats); \Asel separates from \Aproj immediately and the gap widens
monotonically (App.~Fig.~\ref{fig:m5curves}). The full ladder ends at
$1{,}519.0/83.0 = 18.3\times$ lower perplexity than from scratch ($2.91$
nats).

\begin{table}[t]
\centering
\footnotesize
\caption{Seven-arm ablation ladder, primary pair (1.4B$\to$410M), 30M tokens.
Zero-shot is quick-eval perplexity (50k WikiText tokens) at $t{=}0$; final is
full strided WikiText-103 perplexity at budget; here and below a cell is
single-seed unless a $\pm$ appears. Short forms per
App.~Table~\ref{tab:armnames}.}
\label{tab:ladder}
\setlength{\tabcolsep}{3.5pt}
\begin{tabular}{lrr}
\toprule
Arm & Zero-shot ppl & Final ppl (30M) \\
\midrule
\Ascratch & $\sim$66{,}000   & 1{,}519.0 \\
\Aproj    & 12{,}323         & 1{,}054.5 \\
\Ssel     & 11{,}509         & 355.8 \\
\Siso     & 26{,}669         & 208.2 \\
\Scomp    & \textbf{9{,}664} & 114.1 \\
\Srs      & 61{,}912         & 86.4 \\
\Scomprs  & 18{,}459         & \textbf{83.0} \\
\bottomrule
\end{tabular}
\end{table}

\paragraph{Compensation is the function lever.}
\label{sec:race:levers}
Reading the ladder by construction exposes two orthogonal axes over the
selection base: \Acomp re-fits the two purely linear, LayerNorm-free read-out paths ($O$ and
the MLP down-projection) by closed-form ridge least squares
(\S\ref{sec:race:alg}), so surviving units absorb the dropped units' correlated
contribution. This improves both axes: zero-shot $11{,}509\to9{,}664$ (best in
the table) and endpoint $355.8\to114.1$ ($3.1\times$ lower than \Asel, $1.14$
nats).

\paragraph{Rescale is the training-dynamics lever.}
The reference recipe's variance-preserving factor
$\sqrt{d_{\text{in}}/d_{\text{in}}'}$ \citep{samragh2023weight} does the
opposite. On the LayerNorm-free read-out paths only (\Aiso) it worsens
zero-shot $11{,}509\to26{,}669$ yet improves the endpoint $355.8\to208.2$; on
every input-cut path (\Ars) it collapses zero-shot to $61{,}912$ ($5.4\times$
bare selection's $11{,}509$, $1.68$ nats) while driving the endpoint to
$86.4$, past even \Acomp. It does not fix the function (a downstream
LayerNorm cancels the scalar in the forward pass, so it only perturbs
zero-shot alignment) but the variance the optimizer starts from: more
rescale, worse start, better endpoint.

\paragraph{The levers stack.}
The levers act on disjoint quantities, compensation on the function of the two
LN-free read-out paths and rescale on the scale of the LN-fronted read-in paths
($Q$, $K$, $V$, MLP up, output embedding), so they compose: \Acomprs
dominates the strongest single-lever arm (\Ars) on both axes, zero-shot
$18{,}459 < 61{,}912$ and endpoint $83.0 < 86.4$.

\subsection{Compensated Selection}
\label{sec:race:alg}

Algorithm~\ref{alg:compsel} (Appendix~\ref{app:floats}) gives the construction
in pseudocode. \emph{Selection} keeps whole coordinates so each subspace's
symmetry survives: with rotary half-width $16\to8$ the small frequency ladder
is exactly every other large frequency, so rotary pairs are kept at stride $2$
to preserve the $(\cos,\sin)$ pairing \citep{su2024roformer}; when the head
count shrinks (the held-out pair) whole heads shared by $Q,K,V,O$ are kept, and
when depth shrinks, evenly strided blocks with their norms and biases.

\emph{Compensation} is a ridge regression. Let $M$ be a read-out map that
consumes a signal $x$ whose second moment $\Sigma=\mathbb{E}[xx^{\top}]$ we
measure on the frozen corpus, and let $\mathcal{K}$ be the kept input
coordinates. We regress the full map's output $Mx$ onto the kept coordinates
$x_{\mathcal{K}}$:
\begin{equation}
\label{eq:lscomp}
\begin{aligned}
M' &= \arg\min_{M'}\; \mathbb{E}\,\big\lVert Mx - M'x_{\mathcal{K}}\big\rVert^2
      + \lambda\,\lVert M'\rVert_F^2 \\
   &= M\,\Sigma_{:,\mathcal{K}}\,\big(\Sigma_{\mathcal{K},\mathcal{K}} + \lambda I\big)^{-1},
\end{aligned}
\end{equation}
with $\lambda$ scaled to the mean diagonal of $\Sigma_{\mathcal{K},\mathcal{K}}$.
Plain selection is the special case $M'=M_{:,\mathcal{K}}$ (regression
coefficients fixed to the identity on the kept coordinates).
Proposition~\ref{prop:comp} (Appendix~\ref{app:prop})
gives both errors in closed form.

The regression is defined only where its target is: $Mx$ must be a fixed linear
functional of the signal at the cut, which makes a path \emph{well-posed} for
compensation exactly when it is purely linear from the cut to the read, with
no normalization in between. In the Pythia block exactly two paths qualify,
$O$ (reading the concatenated head outputs) and the MLP down-projection
(reading the GELU outputs); every other cut path ($Q$, $K$, $V$, MLP up, the
output embedding) reads the residual through a LayerNorm that renormalizes
over the kept coordinates, so it has no fixed target and takes the rescale.
The placement rule is readable off the architecture
(Figure~\ref{fig:method}a); Observation~\ref{obs:placement}
(Appendix~\ref{app:prop}) states it formally.

\ifappendixfloats\else\AlgCompSel\fi   % deferred to appendix in the page-limited build

\subsection{Hardening}
\label{sec:race:hardening}

\paragraph{Seeded verdict.}
\label{sec:race:seeded}
The two rescale leaders are close on the single seed ($86.4$ vs.\ $83.0$), so we
re-run both across six data-draw seeds: \Acomprs ($84.6\pm2.1$) wins all six
paired comparisons against \Ars ($90.7\pm3.8$; paired $t=7.57$, $p=0.0006$;
App.~Tables~\ref{tab:seeded}, \ref{tab:paired}). Five of its six seeds beat
every \Ars seed; its worst ($88.3$) is above \Ars's best ($86.4$), an
overlap of one seed.
The upset of \Ars ($86.4$) over \Acomp ($114.1$) corrects our earlier reading
that compensation subsumes rescale (Appendix~\ref{app:correction}).

\paragraph{The information-matched control is a designed $2\times2$.}
Compensation alone reads the donor's activation statistics ($\Sigma$ in
Eq.~\ref{eq:lscomp}), so to separate the re-fit from the information it uses
we cross the
\emph{selection statistic} (weight norm, or the expected output energy
$\sqrt{\Sigma_{dd}}\,\lVert\mathrm{col}_d\rVert$ of the informed arms) with
the \emph{re-fit} (whether the read-out maps are compensated) at the same
three data-draw seeds, so comparisons are paired within seed
(Figure~\ref{fig:method}b; App.~Tables~\ref{tab:actsel}
and~\ref{tab:seeded}). On Pythia the re-fit carries the gain: $5.7$
perplexity without the statistic and $7.2$ with it ($3/3$ seeds each) against
$2.3$ for the statistic ($2/3$), and \Acomprs without the statistic beats
\Ainfrs with it in every seed ($84.0$ vs.\ $87.4$); the factors add, and
\Ainfcomprs ($80.2\pm2.3$) is the best Pythia endpoint.

\paragraph{A second architecture family.}
\label{sec:race:family}
Qwen3-1.7B-Base$\to$Qwen3-0.6B-Base (\S\ref{sec:setup}) changes every wiring
detail the decomposition leans on (RMSNorm, SwiGLU, grouped-query attention,
QK-norm, tied embeddings). The placement rule applies unchanged (Figure~\ref{fig:method}a): only the MLP down-projection is
compensated ($\Sigma$ measured at its input, after gating); $Q$, $K$, $V$,
gate, up, and the tied \texttt{lm\_head} (through the final norm's gain) take
the rescale; $O$ reads the uncut attention context, so neither lever touches
it. A bit-exact reconstruction control
($\max|\Delta\,\text{logit}|=0.0$) passed before any race; the protocol is the
primary pair's (30M tokens, three data-draw seeds; details in
Appendix~\ref{app:qwen3}). Three findings transfer
(Table~\ref{tab:qwen3main}). Every rescaled transfer arm ends $10$--$23\times$
lower in perplexity than from scratch at matched budget (Qwen3 \Ascratch
$1{,}217.7\pm30.7$; $2.32$--$3.13$ nats): the best arm $23\times$
($1{,}217.7/53.1$) against $19\times$ on Pythia ($1{,}519.0/80.2$), and
\Acomprs $13.7\times$ ($2.62$ nats) against $18\times$; the re-fit beats plain
selection given the same weight-norm statistic in every seed (\Acomprs
$88.9\pm7.0$ vs.\ \Ars $119.6\pm11.8$, $3/3$, $p=0.010$); and \Ainfcomprs is
the best arm in both families. One finding does not transfer: on Qwen3 the
selection statistic is the big lever ($119.6\to53.4$ without the re-fit,
$88.9\to53.1$ with it, $3/3$ each, $p=0.009$), and the re-fit on top of the
informed selection is a null ($53.4$ vs.\ $53.1$, $+0.3\pm0.6$, $2/3$,
$p=0.48$). The statistic without the re-fit beats the re-fit without the
statistic in every seed ($53.4$ vs.\ $88.9$); on Pythia it was the reverse
($87.4$ vs.\ $84.0$, $3/3$). Two smaller reversals are reported rather than
smoothed: the rescaled arms' zero-shot collapse is far larger than on Pythia
(\Ars $1.0$M and \Acomprs $2.2$M at $t{=}0$, against $61.9$k and $18.5$k),
and the read-out-only rescale (\Aiso) hurts on Qwen3 ($138.5\to249.4$, seed
0) where it helped on Pythia ($355.8\to208.2$).

\paragraph{Why the split flips: the spread of unit energies.}
Per MLP block on both donors we compare the weight-norm score
$\lVert\mathrm{up}_{\mathrm{row}}\rVert\,\lVert\mathrm{down}_{\mathrm{col}}\rVert$
with the energy score $\sqrt{\Sigma_{dd}}\,\lVert\mathrm{down}_{\mathrm{col}}\rVert$
($\Sigma$ on 1{,}000 frozen-corpus sequences; App.~Table~\ref{tab:selstats}).
In both families the two are unrelated (Spearman $-0.07$ on Pythia-1.4B, 24
blocks, and $-0.22$ on Qwen3-1.7B, 28 blocks; top-half overlaps $0.46$ and
$0.42$ against $0.5$ at chance): weight norms carry no information about unit
energy; what differs is the spread of the energies. On
Pythia the coefficient of variation of $\sqrt{\Sigma_{dd}}$ has median $0.24$
over blocks (maximum $2.1$): energies are near-uniform, any half of the units
is nearly as good as any other, and the re-fit does the work. On Qwen3 the
median is $0.62$, with massive-activation outlier
blocks~\citep{sun2024massive} at $50.8$ (block 2) and $6.8$ (block 27): a few
units carry most of the energy, weight norm misses them, informed selection is
the big lever, and the re-fit has little left to repair. Proposition~\ref{prop:comp} says the same in one line:
selection sets the dropped energy
$\operatorname{tr}(M_{\mathcal{D}}\Sigma_{\mathcal{D}\mathcal{D}}M_{\mathcal{D}}^{\top})$,
and the re-fit removes only its linearly predictable part.

\paragraph{Every metric, longer budgets.}
App.~Table~\ref{tab:extended} adds C4 perplexity (out-of-domain), perplexity at
$2\times$ the training context (2048), and four zero-shot tasks: the ordering
\Acomp $<$ \Asel $<$ \Aproj $<$ \Ascratch holds on every metric, and the
multiple-choice tasks sit near guess floors, so 30M tokens buys language
modeling, not task reasoning. At 100M tokens (single seed) \Acomp finishes at
$62.3$ and \Asel at $103.1$, crossing \Asel's 100M endpoint at $\sim$38M
tokens, a $\sim$$2.6\times$ token-efficiency gain (finals are compared only
within a horizon). At 1B tokens ($33\times$ the headline budget) the two
rescale leaders finish level across three data-draw seeds, \Acomprs
$40.3\pm0.3$ vs.\ \Ars $40.3\pm0.5$ (App.~Table~\ref{tab:conv1b}), both
$\sim$$1.4\times$ below from scratch ($57.4$; $0.35$ nats). The parity holds
out of domain (C4 $44.5$ vs.\ $44.4$) and on the four zero-shot tasks; at
$2\times$ context it does not (\Acomprs $74.2\pm4.8$ vs.\ $67.3\pm0.9$, worse
in every seed), which we report rather than smooth. Compensation's advantage
is thus a low-budget effect, redundant with rescale once the budget is large.

\paragraph{A held-out pair.}
The 410M$\to$160M pair is depth-dominated (\S\ref{sec:setup}). Across three
seeds at 30M tokens (\Acomp $113.2\pm4.5$, \Asel $118.5\pm3.3$, \Ascratch
$1{,}505.2\pm57.9$; App.~Table~\ref{tab:unseen}) transfer ends $\sim$$13\times$
below from scratch on every seed, and the full stack keeps its low-budget edge:
\Acomprs ($109.3$) beats \Ars ($117.9$) on all three seeds. Compensation alone
is neutral here (\Acomp and \Asel tie within $\pm1\sigma$) because it repairs
width cuts, not deleted blocks (\S\ref{sec:limitations}); the tie holds at
100M tokens (seed 0: $77.1$ vs.\ $74.4$) and on every other metric
(App.~\ref{app:unseen-metrics}).

\paragraph{Scale, and the method's boundary.}
\label{sec:race:scale}
The scale pair (6.9B$\to$1.4B) reduces all three axes at once at $\sim$5$\times$
the donor size (30M tokens; App.~Table~\ref{tab:scale}). Here the two levers
anti-synergize: each alone beats plain \Asel ($874$; \Ars $572$, \Acomp
$776$), yet their stack \Acomprs ($1{,}213$) is worse than plain selection,
and all still end below from scratch ($1{,}413$), so the core claim survives
its worst case. A control converting the 1.4B donor at
the same reduction ratios did not reproduce the reversal, so the cause appears
to be absolute scale rather than reduction geometry: the compensation solve
(Eq.~\ref{eq:lscomp}) inverts a far larger, worse-conditioned matrix at
$8192$-wide inputs than in the smaller cases ($\le4096$), which its
magnitude-scaled ridge under-regularizes (\S\ref{sec:limitations}).

\subsection{Comparison to knowledge distillation}
\label{sec:race:distill}

Structured pruning followed by knowledge distillation is the standard pipeline
for building a small model from a large one
\citep{muralidharan2024minitron,xia2024sheared}; our selection step is a
structured pruning, so \Ars trained with a distillation loss (the donor 1.4B
as a frozen teacher) is that recipe in our controlled setting (a
matched-budget comparison, not a reproduction of those pipelines). Re-running
the primary pair at 30M tokens over three data-draw seeds with and without
the teacher (App.~Table~\ref{tab:distill}; details in
Appendix~\ref{app:distill}), the two-lever init beats the pruning init in every
seed both without a teacher (\Acomprs $88.2\pm1.2$ vs.\ \Ars $93.9\pm2.0$) and
with one ($80.8\pm2.1$ vs.\ $103.5\pm7.4$), distillation improves \Acomprs in
every seed but degrades the pruning init in every seed, and querying the
teacher every step halves throughput whereas our init reads the donor once.

\section{Conclusion and limitations}
\label{sec:limitations}
\label{sec:conclusion}

\paragraph{Conclusion.}
Within one family, representations align across sizes but parameters do not;
dense projection is destructive because it breaks the architecture's wiring;
and the residuals after the best-fit linear operator carry no exploitable
signal, so conversion value lives in initialization. The placement rule for its two levers transferred
unchanged from Pythia's LayerNorm/GELU block to Qwen3's
RMSNorm/SwiGLU/grouped-query block, and in both families the converted model
ends an order of magnitude below from scratch at matched budget ($19\times$
and $23\times$ for the best arm). Which ingredient carries the gain is a
family property (\S\ref{sec:race:family}); the recommendation of
\S\ref{sec:race:result} is the same in both.

\paragraph{Scope, budgets, and open directions.}
The decomposition leans on which paths a norm fronts (\S\ref{sec:race:alg}),
and the read-out-only rescale changed sign between the families
(\S\ref{sec:race:family}); the Qwen3 evidence is one width-only pair at 30M
tokens on WikiText-103, so more families, depth cuts, and longer budgets there
are future work. Our recovery budgets (30M primary, 100M persistence, 1B
convergence) span $\sim$0.01--0.3\% of Pythia's 300B-token pretraining; at 1B
tokens converted models reach $40.3$ (vs.\ the real 410M's $\approx15.6$), so
most claims concern relative recovery at matched budget rather than end
quality. Open directions: normalization-aware compensation for the
norm-fronted paths (no least-squares target; Observation~\ref{obs:placement})
and reduction-aware conversion for depth cuts (compensation repairs width
cuts, not deleted blocks; \S\ref{sec:race:hardening}).

\paragraph{Compensation conditioning at scale.}
On the 6.9B$\to$1.4B pair (\S\ref{sec:race:scale}) each lever alone beats
plain selection and their stack is worse than plain selection, so the two
levers anti-synergize at this scale; our reading is conditioning of the
Eq.~\ref{eq:lscomp} solve at $8192$-wide inputs, untested remedies are a spectrum-aware ridge, a clamp on the
compensated-weight norm, or importance-based selection, and single-lever \Ars is our default at large donor scale.
   % conclusion and limitations

\subsubsection*{Reproducibility statement}
All experiments use the public Pythia and Qwen3-Base suites and the
\texttt{pile-uncopyrighted}, WikiText-103, and C4 corpora through a seeded,
deterministic pipeline (bit-identical extraction, greedy forward passes, fixed
data order, optimizer, schedule, and budget across arms;
Appendix~\ref{app:protocol}); decisive comparisons report three (headline
pair: six) data-draw seeds; App.~Table~\ref{tab:armnames} maps arm names to
code names. The full code, all configs and seeds, the \texttt{uv.lock}
environment pin, and the frozen-corpus recipe are at
\url{https://github.com/rsdpyenugula/ScaleOp}; the released checkpoints and the
frozen evaluation corpus are at
\url{https://huggingface.co/YenugulaAIML/scaleop-pythia-conversions}.

\subsubsection*{Impact statement}
This paper studies how to initialize a smaller transformer from a larger
pretrained one so that less compute is needed to reach a given quality. The
direct consequence is a lower training cost, and hence lower energy use, for
producing size ladders of language models, which favors groups with modest
compute. A converted model inherits the donor's training data and therefore its
biases and failure modes; conversion adds no filtering or alignment step, so a
deployed converted model needs the same evaluation any pretrained model does.
We release no new data and use only public models and corpora.

\bibliography{references}

\begin{thebibliography}{27}
\providecommand{\natexlab}[1]{#1}
\providecommand{\url}[1]{\texttt{#1}}
\expandafter\ifx\csname urlstyle\endcsname\relax
  \providecommand{\doi}[1]{doi: #1}\else
  \providecommand{\doi}{doi: \begingroup \urlstyle{rm}\Url}\fi

\bibitem[Ainslie et~al.(2023)Ainslie, Lee-Thorp, de~Jong, Zemlyanskiy,
  Lebr{\'o}n, and Sanghai]{ainslie2023gqa}
Joshua Ainslie, James Lee-Thorp, Michiel de~Jong, Yury Zemlyanskiy, Federico
  Lebr{\'o}n, and Sumit Sanghai.
\newblock {GQA}: Training generalized multi-query transformer models from
  multi-head checkpoints.
\newblock In \emph{Proceedings of the 2023 Conference on Empirical Methods in
  Natural Language Processing (EMNLP)}, pp.\  4895--4901, 2023.

\bibitem[Ainsworth et~al.(2023)Ainsworth, Hayase, and
  Srinivasa]{ainsworth2023git}
Samuel~K. Ainsworth, Jonathan Hayase, and Siddhartha Srinivasa.
\newblock Git {Re-Basin}: Merging models modulo permutation symmetries.
\newblock In \emph{International Conference on Learning Representations
  (ICLR)}, 2023.

\bibitem[Ashkboos et~al.(2024)Ashkboos, Croci, Gennari~do Nascimento, Hoefler,
  and Hensman]{ashkboos2024slicegpt}
Saleh Ashkboos, Maximilian~L. Croci, Marcelo Gennari~do Nascimento, Torsten
  Hoefler, and James Hensman.
\newblock {SliceGPT}: Compress large language models by deleting rows and
  columns.
\newblock In \emph{International Conference on Learning Representations
  (ICLR)}, 2024.
\newblock arXiv:2401.15024.

\bibitem[Bansal et~al.(2021)Bansal, Nakkiran, and Barak]{bansal2021revisiting}
Yamini Bansal, Preetum Nakkiran, and Boaz Barak.
\newblock Revisiting model stitching to compare neural representations.
\newblock In \emph{Advances in Neural Information Processing Systems 34
  (NeurIPS)}, 2021.

\bibitem[Biderman et~al.(2023)Biderman, Schoelkopf, Anthony, Bradley, O'Brien,
  Hallahan, Khan, Purohit, Prashanth, Raff, Skowron, Sutawika, and van~der
  Wal]{biderman2023pythia}
Stella Biderman, Hailey Schoelkopf, Quentin Anthony, Herbie Bradley, Kyle
  O'Brien, Eric Hallahan, Mohammad~Aflah Khan, Shivanshu Purohit, USVSN~Sai
  Prashanth, Edward Raff, Aviya Skowron, Lintang Sutawika, and Oskar van~der
  Wal.
\newblock Pythia: A suite for analyzing large language models across training
  and scaling.
\newblock In \emph{Proceedings of the 40th International Conference on Machine
  Learning (ICML)}, pp.\  2397--2430. PMLR, 2023.

\bibitem[Chen et~al.(2022)Chen, Yin, Shang, Jiang, Qin, Wang, Wang, Chen, Liu,
  and Liu]{chen2022bert2bert}
Cheng Chen, Yichun Yin, Lifeng Shang, Xin Jiang, Yujia Qin, Fengyu Wang, Zhi
  Wang, Xiao Chen, Zhiyuan Liu, and Qun Liu.
\newblock {bert2BERT}: Towards reusable pretrained language models.
\newblock In \emph{Proceedings of the 60th Annual Meeting of the Association
  for Computational Linguistics (ACL)}, 2022.

\bibitem[Chen et~al.(2016)Chen, Goodfellow, and Shlens]{chen2016net2net}
Tianqi Chen, Ian Goodfellow, and Jonathon Shlens.
\newblock {Net2Net}: Accelerating learning via knowledge transfer.
\newblock In \emph{International Conference on Learning Representations
  (ICLR)}, 2016.

\bibitem[Davari et~al.(2023)Davari, Horoi, Natik, Lajoie, Wolf, and
  Belilovsky]{davari2023reliability}
MohammadReza Davari, Stefan Horoi, Amine Natik, Guillaume Lajoie, Guy Wolf, and
  Eugene Belilovsky.
\newblock Reliability of {CKA} as a similarity measure in deep learning.
\newblock In \emph{International Conference on Learning Representations
  (ICLR)}, 2023.

\bibitem[Hassibi \& Stork(1993)Hassibi and Stork]{hassibi1993optimal}
Babak Hassibi and David~G. Stork.
\newblock Second order derivatives for network pruning: Optimal brain surgeon.
\newblock In \emph{Advances in Neural Information Processing Systems 5 (NIPS)},
  pp.\  164--171, 1993.

\bibitem[He et~al.(2017)He, Zhang, and Sun]{he2017channel}
Yihui He, Xiangyu Zhang, and Jian Sun.
\newblock Channel pruning for accelerating very deep neural networks.
\newblock In \emph{Proceedings of the IEEE International Conference on Computer
  Vision (ICCV)}, pp.\  1389--1397, 2017.

\bibitem[Hinton et~al.(2015)Hinton, Vinyals, and Dean]{hinton2015distilling}
Geoffrey Hinton, Oriol Vinyals, and Jeff Dean.
\newblock Distilling the knowledge in a neural network.
\newblock \emph{arXiv preprint arXiv:1503.02531}, 2015.

\bibitem[Kornblith et~al.(2019)Kornblith, Norouzi, Lee, and
  Hinton]{kornblith2019similarity}
Simon Kornblith, Mohammad Norouzi, Honglak Lee, and Geoffrey Hinton.
\newblock Similarity of neural network representations revisited.
\newblock In \emph{Proceedings of the 36th International Conference on Machine
  Learning (ICML)}, pp.\  3519--3529. PMLR, 2019.

\bibitem[Lenc \& Vedaldi(2015)Lenc and Vedaldi]{lenc2015understanding}
Karel Lenc and Andrea Vedaldi.
\newblock Understanding image representations by measuring their equivariance
  and equivalence.
\newblock In \emph{Proceedings of the IEEE Conference on Computer Vision and
  Pattern Recognition (CVPR)}, pp.\  991--999, 2015.

\bibitem[Ma et~al.(2023)Ma, Fang, and Wang]{ma2023llmpruner}
Xinyin Ma, Gongfan Fang, and Xinchao Wang.
\newblock {LLM-Pruner}: On the structural pruning of large language models.
\newblock In \emph{Advances in Neural Information Processing Systems 36
  (NeurIPS)}, 2023.

\bibitem[Muralidharan et~al.(2024)Muralidharan, Sreenivas, Joshi, Chochowski,
  Patwary, Shoeybi, Catanzaro, Kautz, and Molchanov]{muralidharan2024minitron}
Saurav Muralidharan, Sharath~Turuvekere Sreenivas, Raviraj Joshi, Marcin
  Chochowski, Mostofa Patwary, Mohammad Shoeybi, Bryan Catanzaro, Jan Kautz,
  and Pavlo Molchanov.
\newblock Compact language models via pruning and knowledge distillation.
\newblock In \emph{Advances in Neural Information Processing Systems
  (NeurIPS)}, 2024.
\newblock arXiv:2407.14679.

\bibitem[Murphy et~al.(2024)Murphy, Zylberberg, and
  Fyshe]{murphy2024correcting}
Alex Murphy, Joel Zylberberg, and Alona Fyshe.
\newblock Correcting biased centered kernel alignment measures in biological
  and artificial neural networks, 2024.
\newblock arXiv:2405.01012.

\bibitem[Samragh et~al.(2023)Samragh, Farajtabar, Mehta, Vemulapalli, Faghri,
  Naik, Tuzel, and Rastegari]{samragh2023weight}
Mohammad Samragh, Mehrdad Farajtabar, Sachin Mehta, Raviteja Vemulapalli,
  Fartash Faghri, Devang Naik, Oncel Tuzel, and Mohammad Rastegari.
\newblock Weight subcloning: direct initialization of transformers using larger
  pretrained ones.
\newblock \emph{arXiv preprint arXiv:2312.09299}, 2023.

\bibitem[Shazeer(2020)]{shazeer2020glu}
Noam Shazeer.
\newblock {GLU} variants improve transformer.
\newblock \emph{arXiv preprint arXiv:2002.05202}, 2020.

\bibitem[Shen et~al.(2022)Shen, Walsh, Keutzer, Dodge, Peters, and
  Beltagy]{shen2022staged}
Sheng Shen, Pete Walsh, Kurt Keutzer, Jesse Dodge, Matthew Peters, and
  Iz~Beltagy.
\newblock Staged training for transformer language models.
\newblock In \emph{Proceedings of the 39th International Conference on Machine
  Learning (ICML)}, pp.\  19893--19908. PMLR, 2022.

\bibitem[Song et~al.(2012)Song, Smola, Gretton, Bedo, and
  Borgwardt]{song2012feature}
Le~Song, Alex Smola, Arthur Gretton, Justin Bedo, and Karsten Borgwardt.
\newblock Feature selection via dependence maximization.
\newblock \emph{Journal of Machine Learning Research}, 13:\penalty0 1393--1434,
  2012.

\bibitem[Su et~al.(2024)Su, Ahmed, Lu, Pan, Bo, and Liu]{su2024roformer}
Jianlin Su, Murtadha Ahmed, Yu~Lu, Shengfeng Pan, Wen Bo, and Yunfeng Liu.
\newblock {RoFormer}: Enhanced transformer with rotary position embedding.
\newblock \emph{Neurocomputing}, 568:\penalty0 127063, 2024.

\bibitem[Sun et~al.(2024)Sun, Chen, Kolter, and Liu]{sun2024massive}
Mingjie Sun, Xinlei Chen, J.~Zico Kolter, and Zhuang Liu.
\newblock Massive activations in large language models.
\newblock In \emph{First Conference on Language Modeling (COLM)}, 2024.
\newblock arXiv:2402.17762.

\bibitem[Tang et~al.(2026)Tang, Wang, Thiele, and Saukh]{tang2026grail}
Wenwu Tang, Dong Wang, Lothar Thiele, and Olga Saukh.
\newblock {GRAIL}: Post-hoc compensation by linear reconstruction for
  compressed networks.
\newblock In Rebekka Burkholz, Shiwei Liu, Saiprasad Ravishankar, William
  Redman, Wei Huang, Weijie Su, and Zhihui Zhu (eds.), \emph{Conference on
  Parsimony and Learning (CPAL)}, volume 328 of \emph{Proceedings of Machine
  Learning Research}, pp.\  881--895. PMLR, 2026.
\newblock URL \url{https://proceedings.mlr.press/v328/tang26a.html}.

\bibitem[Wang et~al.(2023)Wang, Panda, Torroba~Hennigen, Greengard, Karlinsky,
  Feris, Cox, Wang, and Kim]{wang2023ligo}
Peihao Wang, Rameswar Panda, Lucas Torroba~Hennigen, Philip Greengard, Leonid
  Karlinsky, Rogerio Feris, David~D. Cox, Zhangyang Wang, and Yoon Kim.
\newblock Learning to grow pretrained models for efficient transformer
  training.
\newblock In \emph{International Conference on Learning Representations
  (ICLR)}, 2023.

\bibitem[Xia et~al.(2024)Xia, Gao, Zeng, and Chen]{xia2024sheared}
Mengzhou Xia, Tianyu Gao, Zhiyuan Zeng, and Danqi Chen.
\newblock Sheared {LLaMA}: Accelerating language model pre-training via
  structured pruning.
\newblock In \emph{International Conference on Learning Representations
  (ICLR)}, 2024.

\bibitem[Yang et~al.(2025)]{yang2025qwen3}
An~Yang et~al.
\newblock {Qwen3} technical report.
\newblock \emph{arXiv preprint arXiv:2505.09388}, 2025.

\bibitem[Zhang \& Sennrich(2019)Zhang and Sennrich]{zhang2019rmsnorm}
Biao Zhang and Rico Sennrich.
\newblock Root mean square layer normalization.
\newblock In \emph{Advances in Neural Information Processing Systems 32
  (NeurIPS)}, 2019.

\end{thebibliography}
\bibliographystyle{iclr2027_conference}

\clearpage
\appendix
% The appendix tables were laid out for a one-column appendix (6.75in
% text block); widen the block to that width from here on so they fit (arXiv
% has no page geometry constraint).
\setlength{\textwidth}{6.75in}
\setlength{\oddsidemargin}{-0.125in}\setlength{\evensidemargin}{-0.125in}
\setlength{\hsize}{\textwidth}\setlength{\linewidth}{\textwidth}\setlength{\columnwidth}{\textwidth}
\setlength{\headwidth}{\textwidth}
% Appendix floats are [h]; without a barrier the headings pile up while their tables drift pages behind.
\let\origsubsection\subsection
\renewcommand{\subsection}{\FloatBarrier\origsubsection}
\section{Deferred tables and figures}
\label{app:floats}

This appendix collects tables and figures deferred from the main text for space.
All numbers are referenced from, and discussed in, the sections indicated.

\ifappendixfloats\AlgCompSel\TabPredictor\fi

\subsection{Arm names}
\label{app:armnames}

\begin{table}[h]
\centering
\footnotesize
\caption{Arm names. The text names every initialization by its construction;
the released code, Algorithm~\ref{alg:compsel}, and the claims audit use the
code names, and narrow tables use the short form. Every arm is built from the
donor and the frozen corpus alone. ``Rescale'' is the reference recipe's
variance-preserving factor $\sqrt{d_{\text{in}}/d_{\text{in}}'}$
\citep{samragh2023weight}; ``compensation'' is the ridge re-fit of
Eq.~\ref{eq:lscomp}; ``informed'' selection scores the V/O and MLP units by
activation energy $\sqrt{\Sigma_{dd}}\,\lVert\mathrm{col}_d\rVert$ on the
frozen corpus instead of by weight norm (on Qwen3 only the MLP units, since
heads are not cut).}
\label{tab:armnames}
\setlength{\tabcolsep}{4pt}
\begin{tabular}{>{\raggedright\arraybackslash}p{1.55in}>{\raggedright\arraybackslash}p{1.1in}l>{\raggedright\arraybackslash}p{2.45in}}
\toprule
Descriptive name & Short form & Code name & Construction \\
\midrule
\Ascratch & --- & \texttt{random} & target shape at its standard random initialization; no donor \\
\Aproj & --- & \texttt{projection} & dense target-free projection of the donor weights (Eq.~\ref{eq:project}) with neutral norms (gain $1$, bias $0$) \\
\Asel & \Ssel & \texttt{subclone} & whole donor coordinates kept: residual lanes by activation variance; head dims (rotary pairs at stride $2$), heads, and MLP units by weight norm; blocks at even stride \\
\Aiso & \Siso & \texttt{subclone\_iso} & \Asel, then rescale on the norm-free read-outs only ($O$ and MLP down on Pythia; MLP down on Qwen3) \\
\Ars & \Srs & \texttt{subclone\_rs} & \Asel, then rescale on every read whose input is cut: the norm-fronted read-ins ($Q$, $K$, $V$, up, gate, output embedding) and the norm-free read-outs \\
\Acomp & \Scomp & \texttt{hybrid} & \Asel, then ridge compensation of the norm-free read-outs whose input is cut ($O$ where head dims are cut, MLP down) \\
\Acomprs & \Scomprs & \texttt{hybrid\_rs} & \Acomp, then rescale on the norm-fronted read-ins only (Algorithm~\ref{alg:compsel}) \\
\Ainfrs & \Sinfrs & \texttt{subclone\_rs\_act} & \Ars with informed selection \\
\Ainfcomprs & \Sinfcomprs & \texttt{hybrid\_rs\_act} & \Acomprs with informed selection \\
\bottomrule
\end{tabular}
\end{table}

\subsection{Alignment across sizes (\S\ref{sec:representations})}
\label{app:align}

\paragraph{Method.}
Linear CKA, reported as the standard (biased) estimator of
\citet{kornblith2019similarity} with no debiasing, is computed between all
$24\times24$ layer pairs on the similarity subset (500 sequences $\times$ all
token positions; \S\ref{sec:setup}).
The diagonal
mean is $0.883$, yet only $12.5\%$ of layers place their single best match on
the diagonal: the heatmap is one large saturated block (layers 0--2 distinct,
the middle band of layers $\sim$4--22 mutually $\approx1.0$, layer 23
distinct), so linear CKA confirms only coarse early/mid/late correspondence;
saturation in the middle band is a known limitation of the metric.

\paragraph{Calibration.}
The $0.883$ is the biased estimator on the full 64k-token subset. Recomputing
on 16k tokens per layer from the same subset with the biased estimator and
with the debiased one (unbiased HSIC, \citealp{song2012feature}; debiased
CKA, \citealp{murphy2024correcting}) gives the same diagonal mean, $0.898$
and $0.898$, so estimator bias is negligible at this $n$
(Table~\ref{tab:ckacal}). As a reference floor, the same 1.4B activations
against an untrained Pythia-410M (random initialization, same tokens) give
$0.059$ biased and $0.057$ debiased on the diagonal and $0.056$ over all
pairs. The middle-band saturation is therefore real signal, not estimator
bias; the $12.5\%$ diagonal-best fraction, unchanged under debiasing, is why
we call CKA coarse and rest the alignment decisions on held-out ridge $R^2$.
CKA's reliability as a similarity measure is contested more
broadly~\citep{davari2023reliability}, so we use it descriptively; the
alignment claim of \S\ref{sec:representations} rests on the held-out ridge
$R^2=0.84$ of Table~\ref{tab:align} and on behavioral metrics.

\begin{table}[h]
\centering
\footnotesize
\caption{Linear CKA calibration, Pythia-1.4B vs.\ 410M, 16k tokens per layer,
all $24\times24$ layer pairs. ``Diag.'' and ``off-diag.'' are means over the
diagonal and off-diagonal pairs; ``best on diag.'' is the fraction of 1.4B
layers whose best-matching 410M layer has the same index. The untrained rows
are the estimator floor against a randomly initialized 410M.}
\label{tab:ckacal}
\begin{tabular}{llrrrr}
\toprule
410M model & estimator & diag.\ mean & off-diag.\ mean & all pairs & best on diag. \\
\midrule
trained   & biased   & 0.898 & 0.695 & 0.704 & 0.125 \\
trained   & debiased & 0.898 & 0.695 & 0.703 & 0.125 \\
untrained & biased   & 0.059 & 0.056 & 0.056 & 0.042 \\
untrained & debiased & 0.057 & 0.054 & 0.054 & 0.042 \\
\bottomrule
\end{tabular}
\end{table}

This
motivates the directional maps of Table~\ref{tab:align}: per layer, a
large$\,\to\,$small map is fit on pooled activations (8k train / 2k held-out)
under two operators, Procrustes (rotation plus one global scale) and ridge (a
full linear map $2048\!\to\!1024$), and scored by held-out $R^2$. For the
token embeddings the vocabulary rows are split $80/20$ and the same two
operators are fit from the 1.4B rows to the 410M rows.

\begin{figure}[h]
\centering
% source: results/m2_cka/*/cka_heatmap.png and results/m2_maps/*/maps_r2.png
% (gitignored, regenerable at paper quality; dashed reference lines per style guide)
\includegraphics[width=0.46\linewidth]{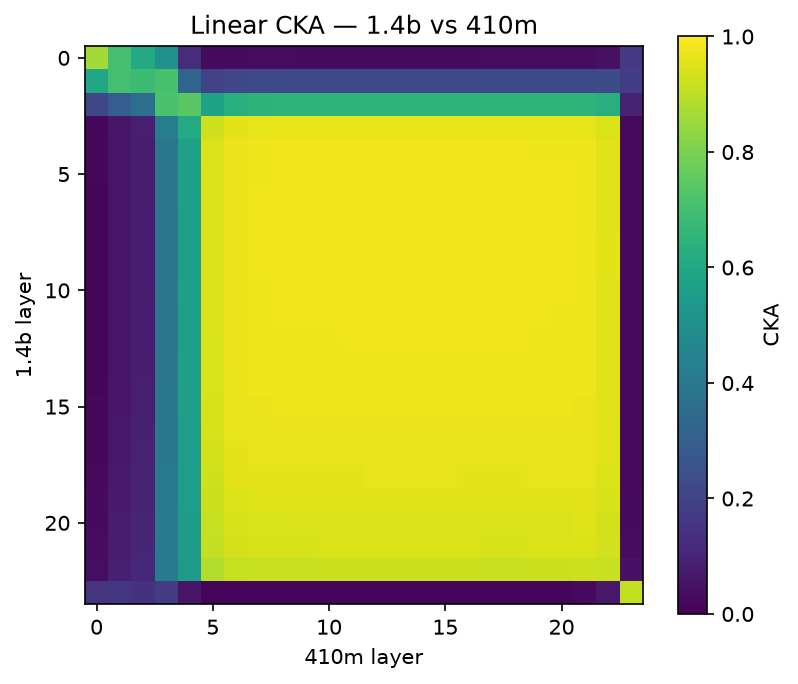}
\hfill
\includegraphics[width=0.46\linewidth]{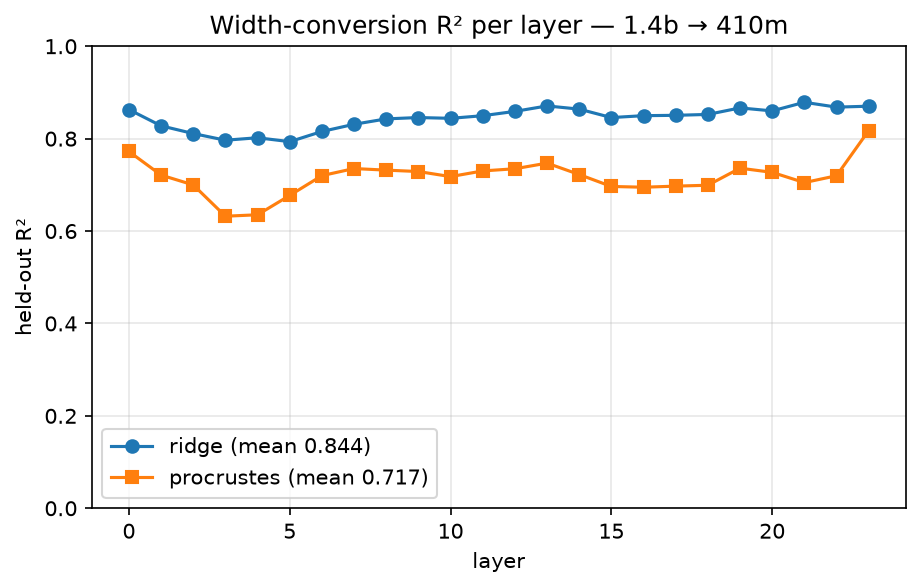}
\caption{Alignment across sizes for the primary pair (1.4B$\,\to\,$410M).
\textbf{(left)} Linear CKA between all $24\times24$ layer pairs: a saturated
middle band (layers $\sim$4--22 mutually $\approx1.0$) confirms coarse
correspondence but resolves no sharp per-layer match. \textbf{(right)} Held-out
$R^2$ of the per-layer activation maps: ridge (full linear) and Procrustes
(rotation$+$scale) track each other and dip together at layers 3--5, the layers
CKA also flags as distinct.}
\label{fig:align}
\end{figure}

\begin{table}[h]
\centering
\caption{Alignment across sizes, held-out. Activations align strongly under a
linear map; raw parameters, even in the cleanest (embedding) case, do not.
Best (highest) map per row in \textbf{bold}.}
\label{tab:align}
\begin{tabular}{lccc}
\toprule
Target & CKA & Ridge $R^2$ & Procrustes $R^2$ \\
\midrule
Pooled activations (per-layer mean) & $0.883^{\dagger}$ & \textbf{0.844} & 0.717 \\
\midrule
Input embedding & 0.512 & \textbf{0.248} & 0.167 \\
Output embedding (unembedding) & 0.678 & \textbf{0.386} & 0.242 \\
\bottomrule
\end{tabular}
\\[3pt]
{\footnotesize $^{\dagger}$ For activations, the diagonal mean of the
$24\times24$ linear-CKA matrix, which saturates in the middle band
(Figure~\ref{fig:align}, left); for embeddings, linear CKA before mapping.}
\end{table}

\subsection{Projection diagnostics (\S\ref{sec:projection})}
\label{app:projection}

\begin{table}[h]
\small
\begin{minipage}[t]{0.55\linewidth}
\centering
\caption{Relative Frobenius error of projected weights vs.\ the actual 410M
weights (mean over 24 layers). Every target-free error exceeds $1$ (worse than
zero); lower per row in \textbf{bold}.}
\label{tab:proj}
\setlength{\tabcolsep}{4pt}
\begin{tabular}{lcc}
\toprule
Weight & Projection & Operator$^{\dagger}$ \\
       & (target-free) & (fitted) \\
\midrule
$Q$            & 1.471 & \textbf{0.658} \\
$K$            & 1.591 & \textbf{0.758} \\
$V$            & 1.390 & \textbf{0.804} \\
$O$            & 1.724 & \textbf{0.795} \\
\texttt{MLP\_UP}   & 1.668 & \textbf{0.683} \\
\texttt{MLP\_DOWN} & 1.723 & \textbf{0.686} \\
\bottomrule
\end{tabular}
\\[3pt]
{\footnotesize $^{\dagger}$ A single $(A,B)$ shared across layers per type; it
peeks at the target and so upper-bounds the linearly explainable fraction. A
\emph{per-layer} fit is degenerate (error $\to 0$ around any full-rank $W$) and
is not reported.}
\end{minipage}\hfill
\begin{minipage}[t]{0.43\linewidth}
\centering
\caption{Zero-shot dense-projected quality (strided WikiText-103 perplexity
before any training); the projected model does not beat random init. Best
(lowest) in \textbf{bold}.}
\label{tab:anchor}
\begin{tabular}{lc}
\toprule
Model & Perplexity \\
\midrule
real 1.4B (reference)        & \textbf{11.5} \\
real 410M (upper anchor)     & 15.6 \\
random-init 410M (lower)     & 65{,}962 \\
\midrule
proj., LN from 410M   & 181{,}242 \\
proj., LN from 1.4B & $10^{13}$ \\
proj., neutral LN$^{\ast}$ & 12{,}323 \\
structured selection$^{\ast}$ & 11{,}509 \\
\bottomrule
\end{tabular}
\\[3pt]
{\footnotesize $^{\ast}$Quick-eval perplexity (the two evals agree at these
magnitudes: random init scores $\approx$66k on both). Neutral norms (gain 1,
bias 0) cut zero-shot perplexity $\sim$15$\times$, so LayerNorm mismatch
dominates the zero-shot collapse; the residual gap to structured selection is
small at $t{=}0$ but $3\times$ in perplexity after training at 30M tokens (the
decomposition in \S\ref{sec:projection}).}
\end{minipage}
\end{table}

\subsection{Spectral residual test (\S\ref{sec:residuals})}
\label{app:spectral}

\begin{table}[h]
\centering
\caption{Test 1 (spectral). Effective rank of the residual $\Delta$ vs.\ a
per-type shuffled control and a shape/scale-matched Gaussian control. $\Delta$
sits $2.8$--$6.4\%$ below both in every type: a faint, non-isotropic
concentration, not a learnable direction.}
\label{tab:erank}
\begin{tabular}{lccc}
\toprule
Weight & $\Delta$ erank & Shuffled & Gaussian \\
\midrule
$Q$            & 771.3 & 824.1 & 824.1 \\
$K$            & 784.0 & 824.0 & 824.1 \\
$V$            & 801.2 & 824.2 & 824.1 \\
$O$            & 792.2 & 824.1 & 824.1 \\
\texttt{MLP\_UP}   & 947.2 & 989.4 & 989.4 \\
\texttt{MLP\_DOWN} & 934.8 & 989.4 & 989.4 \\
\bottomrule
\end{tabular}
\end{table}

\begin{figure}[h]
\centering
% source: results/m4_residuals/*/delta_spectra.png (gitignored, regenerable)
\includegraphics[width=0.6\linewidth]{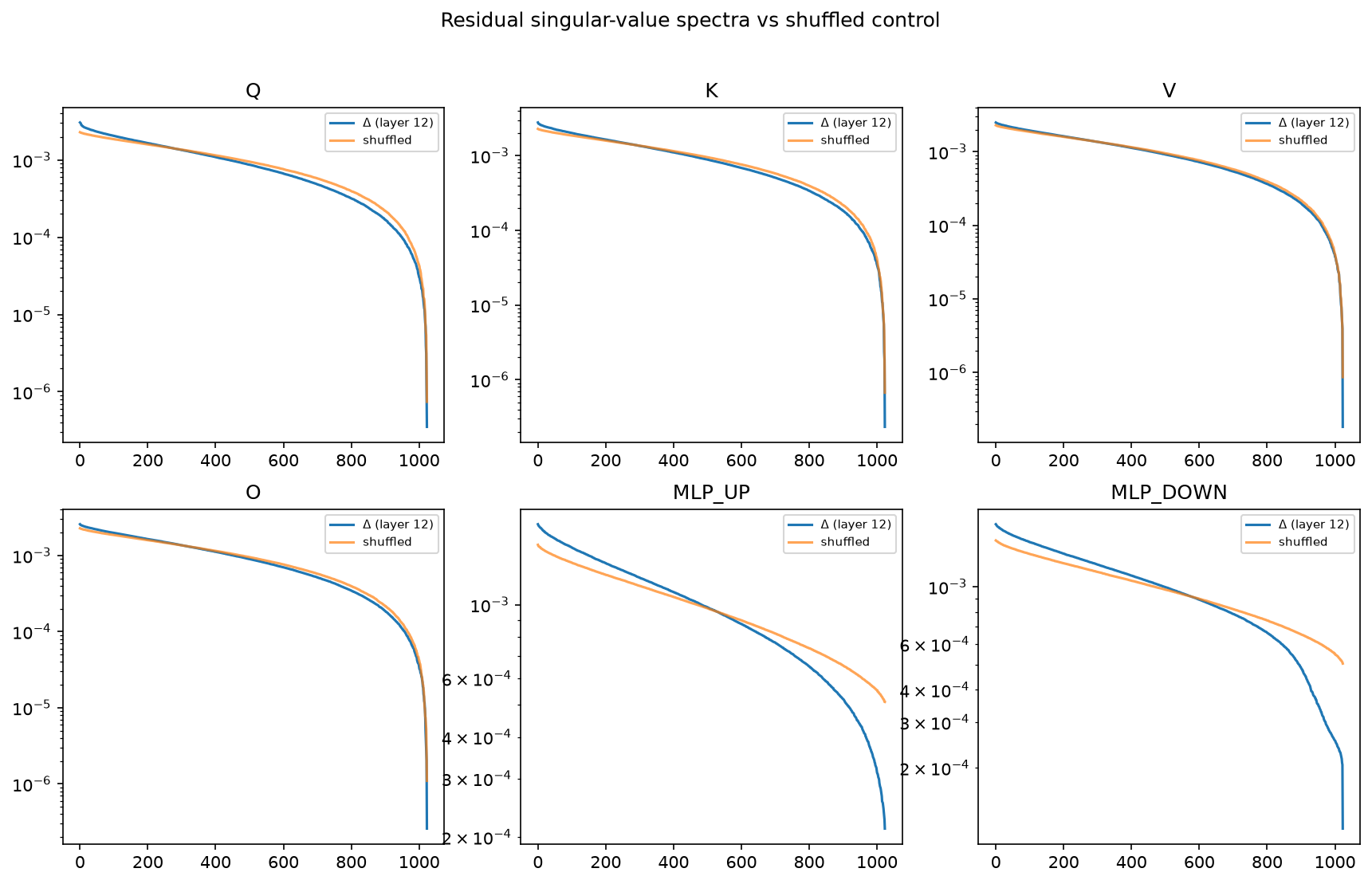}
\caption{Test 1 (spectral). Singular-value spectrum of the residual $\Delta$
against shuffled and Gaussian controls. $\Delta$ concentrates $2.8$--$6.4\%$ more
than either control in every weight type: a faint, generic non-isotropy that
carries no layer-specific, predictable signal (Table~\ref{tab:predictor}).}
\label{fig:spectra}
\end{figure}

\subsection{Extended evaluation, primary pair (\S\ref{sec:race:hardening})}
\label{app:extended}

\begin{table}[h]
\centering
\caption{Extended evaluation of the primary-pair checkpoints after training
at 30M tokens, and the real-410M reference. Guess floors: arc\_easy $0.25$, hellaswag $0.25$, piqa
$0.50$, lambada $\approx0$ (open-vocabulary). Best converted arm in
\textbf{bold}; the ordering holds on all six metrics. Single-seed checkpoints.
This table predates the rescale-lever ablation and reports the compensation
contrast (\Acomp vs.\ \Asel); the rescale gains of
Tables~\ref{tab:ladder}--\ref{tab:seeded} stack on top of it.}
\label{tab:extended}
\begin{tabular}{lrrrrrr}
\toprule
Model & C4 ppl & wk@2048 & lambada & arc\_easy & hellaswag & piqa \\
\midrule
\Acomp & \textbf{90.7} & \textbf{109.8} & \textbf{0.076} & \textbf{0.313} & \textbf{0.266} & \textbf{0.555} \\
\Asel & 162.7 & 346.3 & 0.005 & 0.308 & 0.263 & 0.548 \\
\Aproj & 380.5 & 1{,}210 & 0.000 & 0.272 & 0.258 & 0.539 \\
\Ascratch & 518.8 & 1{,}788 & 0.000 & 0.264 & 0.256 & 0.532 \\
\midrule
real 410M       & 21.0 & 14.5 & 0.516 & 0.519 & 0.337 & 0.667 \\
\bottomrule
\end{tabular}
\end{table}

\subsection{Recovery-ladder curves (\S\ref{sec:race:ladder})}
\label{app:ladder}

\begin{figure}[h]
\centering
\includegraphics[width=0.62\linewidth]{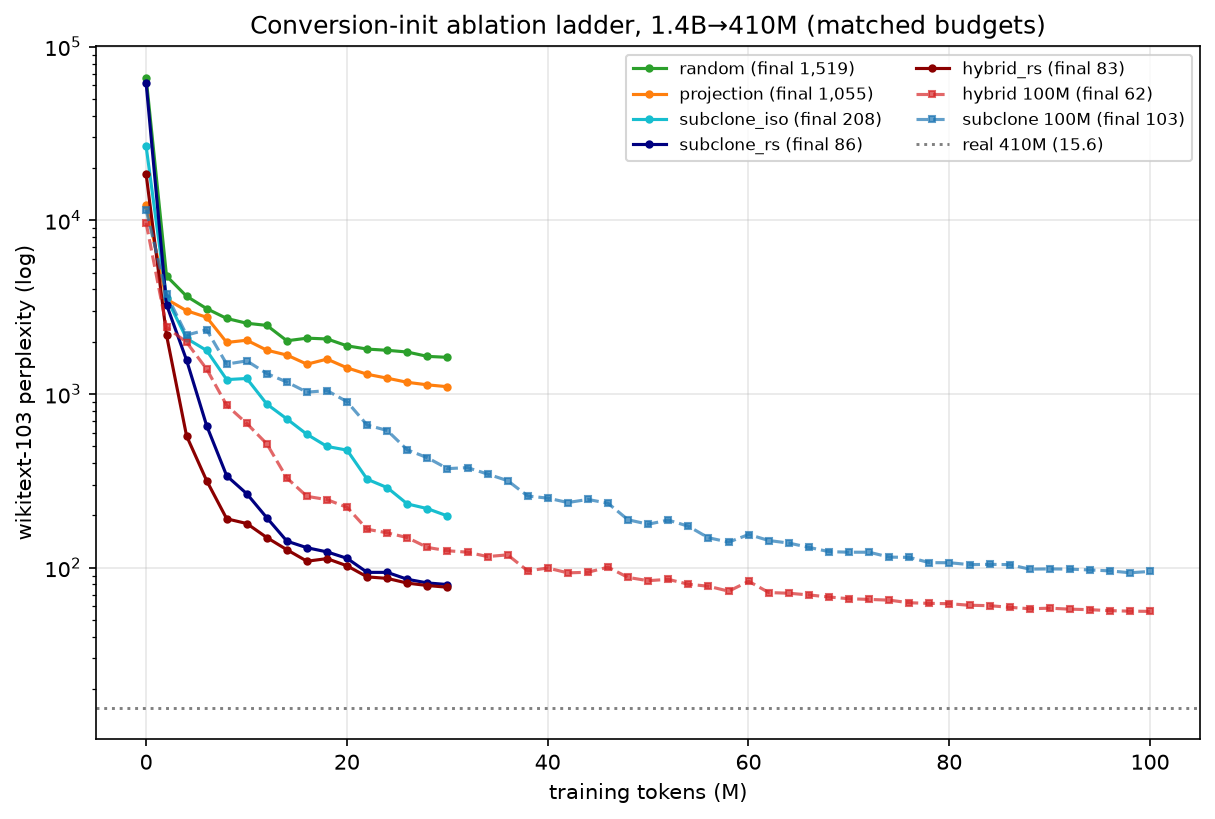}
\caption{Recovery ladder (1.4B$\to$410M). WikiText-103 perplexity (log
scale) vs.\ tokens, seven arms of Table~\ref{tab:ladder}, labeled in the
legend by their code names (Table~\ref{tab:armnames}); dashed $=$ real
Pythia-410M, $\approx15.6$; 30M tokens $=$ $\sim$0.01\% of pretraining, so this
measures relative recovery speed.}
\label{fig:m5curves}
\end{figure}

\subsection{Primary-pair seeded verdict (\S\ref{sec:race:seeded})}
\label{app:seeded}

\begin{table}[h]
\centering
\footnotesize
\caption{Primary-pair final WikiText-103 perplexity across data-draw seeds
(30M tokens; identical budget/schedule per seed). The two headline arms are
run on six seeds: \Acomprs wins every paired comparison, and five of its six
seeds beat every \Ars seed (its seed 3, $88.3$, is above \Ars's seed 0,
$86.4$). The informed-selection rows are the information-matched control of
Table~\ref{tab:actsel}, run on seeds 0--2: V/O and MLP selection scored by
activation energy instead of weight norm.}
\label{tab:seeded}
\begin{tabular}{lrrrrrrr}
\toprule
Init & seed 0 & seed 1 & seed 2 & seed 3 & seed 4 & seed 5 & mean $\pm$ std \\
\midrule
\Ars & 86.4 & 93.7 & 88.9 & 96.3 & 91.3 & 87.5 & $90.7 \pm 3.8$ \\
\Acomprs & \textbf{83.0} & \textbf{86.0} & \textbf{82.8} & \textbf{88.3} & \textbf{84.0} & \textbf{83.4} & $\mathbf{84.6 \pm 2.1}$ \\
\midrule
\Ainfrs & 86.9 & 89.8 & 85.5 & -- & -- & -- & $87.4 \pm 2.2$ \\
\Ainfcomprs & 81.7 & 81.4 & 77.6 & -- & -- & -- & $80.2 \pm 2.3$ \\
\bottomrule
\end{tabular}
\end{table}

\begin{table}[h]
\centering
\scriptsize
\caption{Paired inferential summary of every seeded comparison in the
paper, computed from the printed per-seed final perplexities after training
at the budget named in the first column ($\Delta$ is the first arm
minus the second, so a positive $\Delta$ favors the second, lower-perplexity
arm; ``wins'' counts seeds with $\Delta>0$). The headline row has $n=6$
paired seeds (sign test $6/6$, two-sided $p=0.031$, the smallest value a sign
test can give at $n=6$); every other row has $n=3$, where a sign test cannot
fall below $p=0.25$ two-sided, so ``$3/3$ wins'' is the strongest
nonparametric statement available and the win counts are the primary evidence;
$p$ is a paired $t$-test ($\mathrm{df}=n-1$, two-sided), which assumes normally
distributed differences and is reported as a descriptive complement, not a
gate. Primary-pair rows: the re-fit's effect under each selection statistic
(rows 1--2), the statistic's effect without and with the re-fit (rows 3--4),
and the statistic without the re-fit against the re-fit without the statistic
(row 5). The 1B rows show what a null (WikiText parity) and a reported
negative (perplexity at $2\times$ context) look like on the same footing.
The Qwen3 block mirrors the five primary-pair rows on the second family
(Table~\ref{tab:qwen3seeded}); it and the headline row are computed from the
recorded per-seed values before rounding, so a per-seed $\Delta$ can differ by
$0.1$ from the difference of the printed values. Arms are in the short form of
Table~\ref{tab:armnames}.}
\label{tab:paired}
\setlength{\tabcolsep}{4pt}
\begin{tabular}{llrrrr}
\toprule
Setting & Comparison ($a-b$) & $\Delta$ per seed & mean $\pm$ std & wins & $p$ \\
\midrule
primary, 30M & \Srs $-$ \Scomprs & $+3.4, +7.7, +6.1, +8.0, +7.3, +4.0$ & $+6.1 \pm 2.0$ & 6/6 & $0.0006$ \\
primary, 30M & \Sinfrs $-$ \Sinfcomprs & $+5.2, +8.4, +7.9$ & $+7.2 \pm 1.7$ & 3/3 & $0.019$ \\
primary, 30M & \Srs $-$ \Sinfrs & $-0.5, +3.9, +3.4$ & $+2.3 \pm 2.4$ & 2/3 & $0.245$ \\
primary, 30M & \Scomprs $-$ \Sinfcomprs & $+1.3, +4.6, +5.2$ & $+3.7 \pm 2.1$ & 3/3 & $0.093$ \\
primary, 30M & \Sinfrs $-$ \Scomprs & $+3.9, +3.8, +2.7$ & $+3.5 \pm 0.7$ & 3/3 & $0.012$ \\
\midrule
held-out, 30M & \Srs $-$ \Scomprs & $+6.0, +10.3, +9.6$ & $+8.6 \pm 2.3$ & 3/3 & $0.023$ \\
held-out, 30M & \Ssel $-$ \Scomp & $-2.5, +8.0, +10.4$ & $+5.3 \pm 6.9$ & 2/3 & $0.313$ \\
\midrule
primary, 1B, WikiText & \Srs $-$ \Scomprs & $+0.0, -0.3, +0.4$ & $+0.0 \pm 0.4$ & 1/3 (1 tie) & $0.885$ \\
primary, 1B, ppl at 2048 & \Srs $-$ \Scomprs & $-4.9, -12.9, -2.7$ & $-6.8 \pm 5.4$ & 0/3 & $0.158$ \\
\midrule
distill., batch 16, no teacher & \Srs $-$ \Scomprs & $+3.3, +5.3, +8.3$ & $+5.6 \pm 2.5$ & 3/3 & $0.061$ \\
distill., batch 16, with teacher & \Srs $-$ \Scomprs & $+13.5, +32.3, +22.2$ & $+22.7 \pm 9.4$ & 3/3 & $0.053$ \\
distill., \Scomprs & no teacher $-$ with teacher & $+5.1, +10.3, +6.9$ & $+7.4 \pm 2.6$ & 3/3 & $0.040$ \\
distill., \Srs & no teacher $-$ with teacher & $-5.1, -16.7, -7.0$ & $-9.6 \pm 6.2$ & 0/3 & $0.116$ \\
\midrule
Qwen3, 30M & \Srs $-$ \Scomprs & $+29.7, +36.6, +25.8$ & $+30.7 \pm 5.5$ & 3/3 & $0.010$ \\
Qwen3, 30M & \Sinfrs $-$ \Sinfcomprs & $+0.9, +0.2, -0.3$ & $+0.3 \pm 0.6$ & 2/3 & $0.480$ \\
Qwen3, 30M & \Srs $-$ \Sinfrs & $+60.2, +78.9, +59.6$ & $+66.2 \pm 11.0$ & 3/3 & $0.009$ \\
Qwen3, 30M & \Scomprs $-$ \Sinfcomprs & $+31.4, +42.6, +33.6$ & $+35.9 \pm 5.9$ & 3/3 & $0.009$ \\
Qwen3, 30M & \Sinfrs $-$ \Scomprs & $-30.5, -42.3, -33.8$ & $-35.6 \pm 6.1$ & 0/3 & $0.010$ \\
\bottomrule
\end{tabular}
\end{table}

\subsection{Information-matched control (\S\ref{sec:race:hardening})}
\label{app:actsel}

\begin{table}[h]
\centering
\footnotesize
\caption{Information-matched control as a $2\times2$ design, primary pair
(1.4B$\to$410M), 30M tokens, final WikiText-103 perplexity, mean$\pm$std over
three data-draw seeds (per-seed values in Table~\ref{tab:seeded}; every cell
uses the same three seeds, 0--2, so comparisons are paired within seed; the
six-seed means of the two weight-norm cells are in Table~\ref{tab:seeded}).
Factor 1
(rows): whether the read-out maps are re-fit. Factor 2 (columns): whether V/O
and MLP selection reads the same activation statistics the re-fit uses. Arms
in the short form of Table~\ref{tab:armnames}.}
\label{tab:actsel}
\begin{tabular}{lrr}
\toprule
 & selection by weight norm & selection by activation energy \\
\midrule
no compensation  & $89.7\pm3.7$ (\Srs) & $87.4\pm2.2$ (\Sinfrs) \\
$+$ compensation & $84.0\pm1.8$ (\Scomprs) & $\mathbf{80.2\pm2.3}$ (\Sinfcomprs) \\
\bottomrule
\end{tabular}
\end{table}

\subsection{1B convergence: per-seed finals and extended evaluation (\S\ref{sec:race:hardening})}
\label{app:conv1b}

\begin{table}[h]
\centering
\footnotesize
\caption{Primary pair after training at 1B tokens. Top: final WikiText-103 perplexity per
data-draw seed (seed 0 is the original run; seeds 1--2 re-run on the identical
protocol). Bottom: extended evaluation of the 1B checkpoints, same metrics and
guess floors as Table~\ref{tab:extended}.}
\label{tab:conv1b}
\begin{tabular}{lrrrr}
\toprule
Init & seed 0 & seed 1 & seed 2 & mean $\pm$ std \\
\midrule
\Ars & 40.0 & 40.1 & 40.9 & $40.3 \pm 0.5$ \\
\Acomprs & 40.0 & 40.4 & 40.5 & $40.3 \pm 0.3$ \\
\Ascratch & 57.4 & --   & --   & -- \\
\bottomrule
\end{tabular}
\\[6pt]
\begin{tabular}{llrrrrrr}
\toprule
Init & seed & C4 ppl & wk@2048 & lambada & arc\_easy & hellaswag & piqa \\
\midrule
\Acomprs & 0 & 44.4 & 73.1 & 0.207 & 0.396 & 0.278 & 0.585 \\
\Ars & 0 & 44.4 & 68.2 & 0.209 & 0.392 & 0.276 & 0.588 \\
\Acomprs & 1 & 44.3 & 79.4 & 0.204 & 0.401 & 0.277 & 0.583 \\
\Ars & 1 & 44.3 & 66.5 & 0.208 & 0.400 & 0.277 & 0.587 \\
\Acomprs & 2 & 44.7 & 69.9 & 0.206 & 0.392 & 0.277 & 0.583 \\
\Ars & 2 & 44.5 & 67.2 & 0.205 & 0.394 & 0.276 & 0.594 \\
\midrule
\Acomprs & mean & 44.5 & $74.2\pm4.8$ & 0.206 & 0.396 & 0.277 & 0.584 \\
\Ars & mean & 44.4 & $67.3\pm0.9$ & 0.207 & 0.395 & 0.276 & 0.589 \\
\Ascratch & 0 & 58.5 & 306.2 & 0.128 & 0.355 & 0.268 & 0.557 \\
\midrule
real 410M    & -- & 21.0 & 14.5 & 0.516 & 0.519 & 0.337 & 0.667 \\
\bottomrule
\end{tabular}
\end{table}

\subsection{Held-out pair per-seed detail (\S\ref{sec:race:hardening})}
\label{app:unseen}

\begin{table}[h]
\centering
\caption{Held-out pair (410M$\to$160M), final WikiText-103 perplexity across
three data-draw seeds (30M tokens). Transfer ends $\sim$$13\times$ below from
scratch in perplexity on every seed; \Acomp and \Asel are a statistical tie
on this depth-dominated pair (overlapping $\pm1\sigma$); the full stack keeps
its edge, \Acomprs beating \Ars on every seed.}
\label{tab:unseen}
\begin{tabular}{lrrrr}
\toprule
Init & seed 0 & seed 1 & seed 2 & mean $\pm$ std \\
\midrule
\Acomp & 117.5 & 113.6 & 108.5 & $\mathbf{113.2 \pm 4.5}$ \\
\Asel & 115.0 & 121.6 & 118.9 & $118.5 \pm 3.3$ \\
\Acomprs & \textbf{111.3} & \textbf{111.1} & \textbf{105.4} & $\mathbf{109.3 \pm 3.4}$ \\
\Ars & 117.3 & 121.4 & 115.0 & $117.9 \pm 3.2$ \\
\Ascratch & 1{,}464.4 & 1{,}571.5 & 1{,}479.6 & $1{,}505.2 \pm 57.9$ \\
\bottomrule
\end{tabular}
\end{table}

\paragraph{Metric robustness (held-out pair).}
\label{app:unseen-metrics}
On the held-out 410M$\to$160M pair the tie between \Acomp and \Asel holds on
every metric beyond WikiText-103: C4 perplexity $91.1$ vs.\ $94.2$, perplexity at
$2\times$ context $111.8$ vs.\ $110.2$, LAMBADA $11.0\%$ vs.\ $10.7\%$ (others
within noise), while \Ascratch sits at C4 $514$ and LAMBADA $0$; the real-160M
anchor (LAMBADA $35.4\%$) matches published Pythia numbers. These cells are
single-seed.

\begin{figure}[h]
\centering
\includegraphics[width=0.92\linewidth]{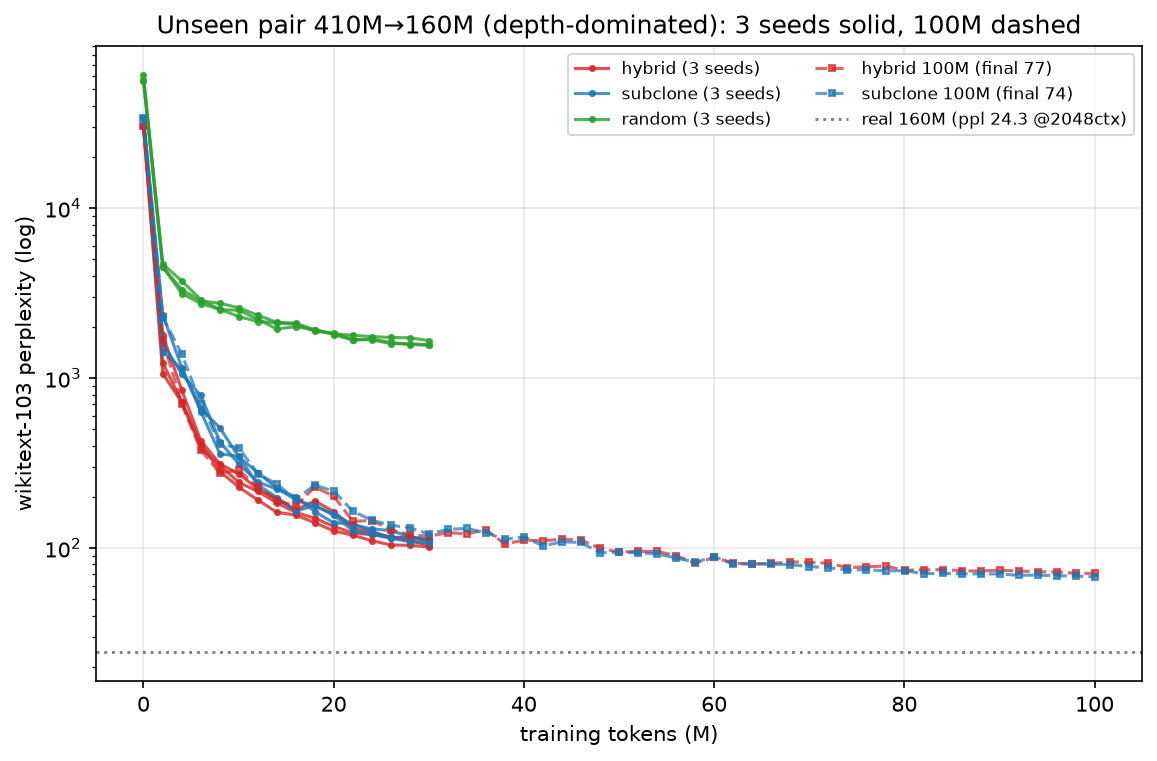}
\caption{\textbf{Held-out pair (410M$\to$160M), three seeds.} Full WikiText-103
perplexity (log scale) versus tokens; bands span the three data-draw seeds and
the dashed line marks the real Pythia-160M reference. Transfer inits (\Acomp,
\Asel; the legend uses the code names of Table~\ref{tab:armnames}) sit an
order of magnitude below \Ascratch throughout and overlap each other, the depth-dominated regime where compensation's width-repair edge is
neutral (Table~\ref{tab:unseen}), yet the $\sim$$13\times$ perplexity margin
over from scratch is preserved on every seed.}
\label{fig:m5bcurves}
\end{figure}

\subsection{Scale pair (\S\ref{sec:race:scale})}
\label{app:scale}

\begin{table}[h]
\centering
\caption{Scale pair (6.9B$\to$1.4B): final WikiText-103 perplexity after
training at 30M tokens, single seed. Each lever alone beats plain selection
and their stack is worse than plain selection, so the two levers
anti-synergize at this scale; every transfer arm still ends below from scratch
(\S\ref{sec:race:scale}).}
\label{tab:scale}
\begin{tabular}{lr}
\toprule
Arm & Final ppl (30M) \\
\midrule
\Ascratch & 1{,}413 \\
\Asel     & 874 \\
\Acomp    & 776 \\
\Ars      & \textbf{572} \\
\Acomprs  & 1{,}213 \\
\bottomrule
\end{tabular}
\end{table}

\subsection{Distillation comparison (\S\ref{sec:race:distill})}
\label{app:distill}

\paragraph{Setup.}
The distillation loss is
$L=\tfrac12\,\mathrm{CE}+\tfrac12\,T^{2}\,\mathrm{KL}(\text{student}\,\Vert\,\text{teacher})$
with the donor 1.4B as a frozen teacher and $T{=}2$. The teacher forward caps
the batch at 16 sequences, so the no-distillation arms are re-run at batch 16
as well and the comparison is matched. Three findings hold in all three seeds
(3/3 paired wins in both columns of Table~\ref{tab:distill}; paired tests in
Table~\ref{tab:paired}): the two-lever init beats the pruning init with and
without a teacher; distillation improves \Acomprs in every seed
($88.2\to80.8$ on average) but degrades the pruning init in every seed
($93.9\to103.5$), and \Acomprs with distillation is the best arm in
every seed; and querying the teacher every step halves throughput
($\sim$4.5k vs.\ $\sim$9.4k tok/s on the GB10), whereas the two-lever init
reads the donor once at construction.

\begin{table}[h]
\centering
\footnotesize
\caption{Adding knowledge distillation (frozen 1.4B teacher) to continued
pre-training, primary pair (1.4B$\to$410M), 30M tokens, batch 16 throughout for a
matched comparison. Final full WikiText-103 perplexity, mean$\pm$std over three
data-draw seeds (\Ascratch: single seed); \Ars is a
structured-pruning init, so its distilled column is the prune-plus-distill
recipe.}
\label{tab:distill}
\begin{tabular}{lrr}
\toprule
Init & no distillation & $+$ distillation \\
\midrule
\Ascratch & 1{,}359.4 & 1{,}311.3 \\
\Ars (pruning) & $93.9\pm2.0$ & $103.5\pm7.4$ \\
\Acomprs (ours) & $\mathbf{88.2\pm1.2}$ & $\mathbf{80.8\pm2.1}$ \\
\bottomrule
\end{tabular}
\end{table}

\subsection{Second family: Qwen3-1.7B-Base$\to$Qwen3-0.6B-Base (\S\ref{sec:race:family})}
\label{app:qwen3}

\paragraph{Pair and placement.}
Both models have 28 blocks, grouped-query attention with 16 query and 8
key-value heads at head dimension 128, QK-norm, RMSNorm, gated SwiGLU, tied
input/output embeddings, no attention biases, and the same tokenizer; the cut
is width-only, hidden $2048\to1024$ and MLP intermediate $6144\to3072$. Heads
and head dimension do not change, so there is no head, rotary, or V/O
selection; the two cut axes are the residual stream (scored by the variance of
each block's mean-pooled output over the frozen corpus, averaged over blocks,
as on Pythia) and the MLP intermediate (weight norm
$\lVert\mathrm{up}_{\text{row}}\rVert\,\lVert\mathrm{down}_{\text{col}}\rVert$,
or activation energy $\sqrt{\Sigma_{dd}}\,\lVert\mathrm{down}_{\text{col}}\rVert$
in the informed-selection arms). The placement rule of \S\ref{sec:race:alg} reads
off the wiring as on Pythia: $Q$, $K$, $V$, gate, and up read the residual
through RMSNorm, which renormalizes over the kept coordinates exactly as
LayerNorm does, so they take the rescale lever only; the tied output embedding
reads the final RMSNorm's output, and with tied weights its rescale is applied
through the final norm's gain (the forward pass is identical); the MLP
down-projection reads $\mathrm{silu}(\text{gate})\odot\text{up}$ with no norm
in between and takes ridge compensation (Eq.~\ref{eq:lscomp}) with $\Sigma$
measured at its input, after gating, on 1{,}000 frozen-corpus sequences; $O$
reads the uncut attention context, so its input takes neither lever and only
its output rows follow the residual selection. Rebuilding the donor from its
own extracted roles reproduces its logits with $\max|\Delta\,\text{logit}|=0.0$
(the reconstruction control of \S\ref{sec:setup}), checked before any race.

\paragraph{Protocol.}
Budget, schedule, optimizer, and data protocol are the primary pair's (30M
tokens, batch $32\times1024$, learning rate $3\times10^{-4}$, 100 warmup
steps, cosine decay, three data-draw seeds with the same stream offsets;
Appendix~\ref{app:protocol}), with the frozen corpus re-tokenized under the
Qwen3 tokenizer and the same contamination guard. Models instantiated from the
hub configuration default to bf16 parameters, so the from-scratch arm is forced
to fp32 master weights to match the transfer arms (all arms train under bf16
autocast). Zero-shot is quick-eval perplexity at $t{=}0$; final is full
strided WikiText-103 perplexity at budget.

\begin{table}[h]
\centering
\footnotesize
\caption{Qwen3 ladder, seed 0, 30M tokens. Zero-shot is quick-eval perplexity
at $t{=}0$; final is full WikiText-103 perplexity at budget. The read-out-only
rescale (\Aiso) hurts here where it helped on Pythia
(Table~\ref{tab:ladder}), and the rescaled arms' zero-shot collapse is larger
than on Pythia; both are reported, not smoothed.}
\label{tab:qwen3ladder}
\begin{tabular}{lrr}
\toprule
Arm & Zero-shot ppl & Final ppl (30M) \\
\midrule
\Ascratch (fp32) & 180{,}131   & 1{,}183.7 \\
\Asel            & 22{,}740    & 138.5 \\
\Aiso            & 27{,}422    & 249.4 \\
\Acomp           & \textbf{11{,}289} & 89.4 \\
\Ars             & 1{,}025{,}033 & 112.8 \\
\Acomprs         & 2{,}173{,}995 & 83.1 \\
\Ainfrs          & 114{,}028   & 52.6 \\
\Ainfcomprs      & 33{,}030 & \textbf{51.6} \\
\bottomrule
\end{tabular}
\end{table}

\begin{table}[h]
\centering
\footnotesize
\caption{Qwen3 pair, final WikiText-103 perplexity across three data-draw
seeds (30M tokens). Every arm uses the same seeds and protocol, so comparisons
are paired within seed (Table~\ref{tab:paired}, Qwen3 block). Best mean in
\textbf{bold}.}
\label{tab:qwen3seeded}
\begin{tabular}{lrrrr}
\toprule
Init & seed 0 & seed 1 & seed 2 & mean $\pm$ std \\
\midrule
\Ascratch (fp32) & 1{,}183.7 & 1{,}243.3 & 1{,}226.3 & $1{,}217.7 \pm 30.7$ \\
\Ars & 112.8 & 133.2 & 112.9 & $119.6 \pm 11.8$ \\
\Acomprs & 83.1  & 96.7  & 87.1  & $88.9 \pm 7.0$ \\
\Ainfrs & 52.6  & 54.3  & 53.3  & $53.4 \pm 0.9$ \\
\Ainfcomprs & 51.6  & 54.1  & 53.5  & $\mathbf{53.1 \pm 1.3}$ \\
\bottomrule
\end{tabular}
\end{table}

\begin{table}[h]
\centering
\footnotesize
\caption{Information-matched $2\times2$ on the Qwen3 pair (companion to
Table~\ref{tab:actsel}), 30M tokens, final WikiText-103 perplexity,
mean$\pm$std over three data-draw seeds (per-seed values in
Table~\ref{tab:qwen3seeded}). Here the column factor (selection statistic)
carries the gain and the row factor (re-fit) is a null in the energy column.
Arms in the short form of Table~\ref{tab:armnames}.}
\label{tab:qwen3actsel}
\begin{tabular}{lrr}
\toprule
 & selection by weight norm & selection by activation energy \\
\midrule
no compensation  & $119.6\pm11.8$ (\Srs) & $53.4\pm0.9$ (\Sinfrs) \\
$+$ compensation & $88.9\pm7.0$ (\Scomprs) & $\mathbf{53.1\pm1.3}$ (\Sinfcomprs) \\
\bottomrule
\end{tabular}
\end{table}

\begin{table}[h]
\centering
\footnotesize
\caption{Selection-statistic diagnostic on the two donors
(\S\ref{sec:race:family}). Per MLP block, every intermediate unit is scored by
weight norm $\lVert\mathrm{up}_{\text{row}}\rVert\,\lVert\mathrm{down}_{\text{col}}\rVert$
and by activation energy $\sqrt{\Sigma_{dd}}\,\lVert\mathrm{down}_{\text{col}}\rVert$
($\Sigma$ at the down-projection's input on 1{,}000 frozen-corpus sequences).
Columns: Spearman correlation of the two scores (mean over blocks), overlap of
the two top-half sets ($0.5$ is chance), median and maximum over blocks of the
coefficient of variation of $\sqrt{\Sigma_{dd}}$ across units (block index in
parentheses), and the number of blocks with Spearman below $-0.2$.}
\label{tab:selstats}
\begin{tabular}{lrrrrlr}
\toprule
Donor & blocks & Spearman & top-half overlap & median CV & max CV (block) & blocks $<-0.2$ \\
\midrule
Pythia-1.4B      & 24 & $-0.07$ & 0.46 & 0.24 & 2.1 (3)             & 5 \\
Qwen3-1.7B-Base  & 28 & $-0.22$ & 0.42 & 0.62 & 50.8 (2); 6.8 (27)  & 19 \\
\bottomrule
\end{tabular}
\end{table}

\begin{table}[h]
\centering
\caption{Extended evaluation of the Qwen3 seed-0 checkpoints after training at
30M tokens, and the real Qwen3-0.6B-Base reference, the companion of Table~\ref{tab:extended}
(same metrics and guess floors; fp32 weights throughout). Best converted arm
per column in \textbf{bold}: the two informed-selection arms split the best
cells and lead every metric, including perplexity at $2\times$ the training
context, where this family shows none of the penalty seen for the Pythia
\Acomprs at 1B tokens (Table~\ref{tab:conv1b}). Every transfer
arm beats from scratch on every metric; multiple-choice tasks sit near their
guess floors at this budget, as on Pythia. Single-seed checkpoints.}
\label{tab:qwen3extended}
\footnotesize
\setlength{\tabcolsep}{4pt}
\begin{tabular}{lrrrrrr}
\toprule
Model & C4 ppl & wk@2048 & lambada & arc\_easy & hellaswag & piqa \\
\midrule
\Ainfcomprs & \textbf{62.0} & \textbf{48.1} & 0.137 & 0.400 & 0.272 & 0.575 \\
\Ainfrs & 62.6 & 49.3 & \textbf{0.138} & \textbf{0.406} & \textbf{0.274} & \textbf{0.579} \\
\Acomprs & 85.4 & 76.1 & 0.104 & 0.355 & 0.265 & 0.568 \\
\Ars & 104.8 & 104.7 & 0.078 & 0.364 & 0.267 & 0.565 \\
\Acomp & 88.5 & 82.8 & 0.104 & 0.359 & 0.264 & 0.577 \\
\Asel & 119.5 & 125.8 & 0.061 & 0.341 & 0.266 & 0.562 \\
\Aiso & 185.0 & 230.3 & 0.013 & 0.321 & 0.260 & 0.550 \\
\Ascratch & 616.8 & 1{,}164 & 0.000 & 0.269 & 0.260 & 0.534 \\
\midrule
real Qwen3-0.6B-Base & 19.8 & 11.5 & 0.543 & 0.653 & 0.410 & 0.702 \\
\bottomrule
\end{tabular}
\end{table}

\section{Deferred discussion}
\label{app:discussion}

\subsection{What compensation removes (\S\ref{sec:race:alg}, \S\ref{sec:race:family})}
\label{app:prop}

Let $M\in\mathbb{R}^{m\times n}$ be a linear read-out of a signal
$x\in\mathbb{R}^{n}$ with second moment $\Sigma=\mathbb{E}[xx^{\top}]$, let a
cut keep the coordinates $\mathcal{K}$ and drop $\mathcal{D}$, and write
$Mx = M_{\mathcal{K}}x_{\mathcal{K}} + M_{\mathcal{D}}x_{\mathcal{D}}$ with
$M_{\mathcal{K}}=M_{:,\mathcal{K}}$, $M_{\mathcal{D}}=M_{:,\mathcal{D}}$, and
block indices abbreviated ($\Sigma_{\mathcal{KK}}=\Sigma_{\mathcal{K},\mathcal{K}}$,
$\Sigma_{\mathcal{DK}}=\Sigma_{\mathcal{D},\mathcal{K}}$, and so on). A
substitute map $M'$ acting on $x_{\mathcal{K}}$ has expected squared error
$\varepsilon(M') = \mathbb{E}\lVert Mx - M'x_{\mathcal{K}}\rVert^{2}$.

\begin{proposition}[Compensation removes the predictable part of the cut]
\label{prop:comp}
Assume $\Sigma_{\mathcal{KK}}\succ0$.
\begin{enumerate}
\item[(i)] Plain selection, $M'=M_{\mathcal{K}}$, has
  $\varepsilon(M_{\mathcal{K}}) = \operatorname{tr}\!\big(M_{\mathcal{D}}\Sigma_{\mathcal{DD}}M_{\mathcal{D}}^{\top}\big)$,
  the output energy the dropped coordinates carried.
\item[(ii)] The least-squares map
  $M^{\star} = M\,\Sigma_{:,\mathcal{K}}\,\Sigma_{\mathcal{KK}}^{-1}$, the
  $\lambda\to0$ limit of Eq.~\ref{eq:lscomp}, is the unique minimizer of
  $\varepsilon$, equals
  $M_{\mathcal{K}} + M_{\mathcal{D}}\Sigma_{\mathcal{DK}}\Sigma_{\mathcal{KK}}^{-1}$,
  and has
  $\varepsilon(M^{\star}) = \operatorname{tr}\!\big(M_{\mathcal{D}}\Sigma_{\mathcal{D}|\mathcal{K}}M_{\mathcal{D}}^{\top}\big)$
  with
  $\Sigma_{\mathcal{D}|\mathcal{K}} = \Sigma_{\mathcal{DD}} - \Sigma_{\mathcal{DK}}\Sigma_{\mathcal{KK}}^{-1}\Sigma_{\mathcal{KD}}$.
  Hence
  $\varepsilon(M_{\mathcal{K}}) - \varepsilon(M^{\star})
   = \operatorname{tr}\!\big(M_{\mathcal{D}}\Sigma_{\mathcal{DK}}\Sigma_{\mathcal{KK}}^{-1}\Sigma_{\mathcal{KD}}M_{\mathcal{D}}^{\top}\big) \ge 0$:
  compensation removes exactly the part of the dropped coordinates that is
  linearly predictable from the kept ones, with equality iff
  $M_{\mathcal{D}}\Sigma_{\mathcal{DK}}=0$, in particular whenever the dropped
  coordinates are uncorrelated with the kept ones ($\Sigma_{\mathcal{DK}}=0$).
\item[(iii)] For $\lambda>0$ the ridge objective of Eq.~\ref{eq:lscomp} is a
  strictly convex quadratic in $M'$ (for any $\Sigma_{\mathcal{KK}}\succeq0$),
  and $M\,\Sigma_{:,\mathcal{K}}(\Sigma_{\mathcal{KK}}+\lambda I)^{-1}$ is its
  unique minimizer.
\end{enumerate}
\end{proposition}

\begin{proof}
(i) $Mx - M_{\mathcal{K}}x_{\mathcal{K}} = M_{\mathcal{D}}x_{\mathcal{D}}$, so
$\varepsilon(M_{\mathcal{K}})
 = \mathbb{E}\lVert M_{\mathcal{D}}x_{\mathcal{D}}\rVert^{2}
 = \operatorname{tr}\big(M_{\mathcal{D}}\,\mathbb{E}[x_{\mathcal{D}}x_{\mathcal{D}}^{\top}]\,M_{\mathcal{D}}^{\top}\big)$.
(ii) Expanding,
$\varepsilon(M') = \mathbb{E}\lVert Mx\rVert^{2}
 - 2\operatorname{tr}(M'\Sigma_{\mathcal{K},:}M^{\top})
 + \operatorname{tr}(M'\Sigma_{\mathcal{KK}}M'^{\top})$,
a quadratic in $M'$ whose Hessian is $\Sigma_{\mathcal{KK}}\otimes I\succ0$,
so it is strictly convex with the unique stationary point
$M'\Sigma_{\mathcal{KK}} = M\Sigma_{:,\mathcal{K}}$, i.e.\ $M^{\star}$. Since
$M\Sigma_{:,\mathcal{K}} = M_{\mathcal{K}}\Sigma_{\mathcal{KK}} + M_{\mathcal{D}}\Sigma_{\mathcal{DK}}$,
$M^{\star} = M_{\mathcal{K}} + M_{\mathcal{D}}\Sigma_{\mathcal{DK}}\Sigma_{\mathcal{KK}}^{-1}$:
plain selection plus the best linear prediction of the dropped coordinates
from the kept ones, routed through $M_{\mathcal{D}}$. Then
$Mx - M^{\star}x_{\mathcal{K}} = M_{\mathcal{D}}\,r$ with
$r = x_{\mathcal{D}} - \Sigma_{\mathcal{DK}}\Sigma_{\mathcal{KK}}^{-1}x_{\mathcal{K}}$,
the residual of that prediction, and
$\mathbb{E}[rr^{\top}] = \Sigma_{\mathcal{D}|\mathcal{K}}$ by direct
expansion, which gives $\varepsilon(M^{\star})$. Subtracting, the gap is
$\operatorname{tr}\big(M_{\mathcal{D}}\Sigma_{\mathcal{DK}}\Sigma_{\mathcal{KK}}^{-1}\Sigma_{\mathcal{KD}}M_{\mathcal{D}}^{\top}\big)
 = \big\lVert\Sigma_{\mathcal{KK}}^{-1/2}\Sigma_{\mathcal{KD}}M_{\mathcal{D}}^{\top}\big\rVert_{F}^{2}\ge0$,
zero iff $\Sigma_{\mathcal{KD}}M_{\mathcal{D}}^{\top}=0$.
(iii) Adding $\lambda\lVert M'\rVert_{F}^{2}$ adds $\lambda I$ to the Hessian,
which is then positive definite for any $\Sigma_{\mathcal{KK}}\succeq0$; the
normal equations $M'(\Sigma_{\mathcal{KK}}+\lambda I) = M\Sigma_{:,\mathcal{K}}$
give the closed form.
\end{proof}

The placement rule of \S\ref{sec:race:alg} and Figure~\ref{fig:method}a, in
formal form:

\begin{observation}[Placement rule, in this wiring]
\label{obs:placement}
Let a cut remove input coordinates of a map $M$, keeping $\mathcal{K}$. If the
path from the cut to $M$ is linear (no normalization between them), the
compensated map of Eq.~\ref{eq:lscomp} is the unique ridge-optimal substitute
for $M$ under the measured second moment $\Sigma$, and plain selection is its
special case. If a LayerNorm sits between the cut and $M$, the pre-cut read is
not a linear function of $x_{\mathcal{K}}$, so no such regression target
exists.
\end{observation}
\noindent\emph{Justification.} The first case is Proposition~\ref{prop:comp}
(for $\lambda>0$ the objective is a strictly convex quadratic whose normal
equations give the closed form); in the second, $M$ receives
$\mathrm{LN}(x_{\mathcal{K}})$, whose mean and variance are recomputed over
$\mathcal{K}$, so $M\,\mathrm{LN}(x)$ is not linear in $x_{\mathcal{K}}$.

Read against the $2\times2$ of \S\ref{sec:race:family}: informed selection
lowers $\operatorname{tr}(M_{\mathcal{D}}\Sigma_{\mathcal{DD}}M_{\mathcal{D}}^{\top})$
directly by placing low-energy units in $\mathcal{D}$, which is why it
dominates when unit energies are heterogeneous (Qwen3,
Table~\ref{tab:selstats}), while the re-fit recovers only the predictable
part,
$\operatorname{tr}(M_{\mathcal{D}}\Sigma_{\mathcal{DK}}\Sigma_{\mathcal{KK}}^{-1}\Sigma_{\mathcal{KD}}M_{\mathcal{D}}^{\top})$,
which is what remains to be gained when energies are near-uniform and any half
carries about the same energy (Pythia). The second case of
Observation~\ref{obs:placement} is unchanged: with a normalization between the
cut and $M$ there is no fixed linear target and the decomposition does not
apply.

\subsection{An honest correction (\S\ref{sec:race:seeded})}
\label{app:correction}

The rescale lever corrects our own earlier reading. When \Acomp first won at
$114.1$ we argued that least-squares compensation \emph{subsumed} the reference
recipe's scalar rescale, since a ridge-optimal map beats a scalar on the paths it
re-fits. The upset (\Ars, with no compensation, at $86.4$ against \Acomp)
showed that is only half right: compensation dominates rescale
\emph{on the two read-out paths it touches}, but the LN-fronted read-in paths it
leaves alone still benefit from the reference recipe's variance correction as a
training-dynamics prior. Blanket rescale collapses zero-shot yet wins on
dynamics, so the reference recipe is vindicated exactly where our first account
wrote it off, and the principled method is the stack, not either lever alone.

\subsection{Reconciling the faint spectral signal (\S\ref{sec:residuals})}
\label{app:reconcile}

Tests 1 and 3 are not in tension: the $2.8$--$6.4\%$ spectral deficit is a
\emph{generic} statistical trace (mild row/column-norm heterogeneity), not a
correspondence between $\hat{W}$ and $\Delta$ any predictor can exploit. A
behavioral cross-check agrees: an assembled 410M whose six matrix types are
operator-projected (with real biases, LayerNorms, and embeddings) scores
$35{,}212$ perplexity, and the learned correction moves it only $\sim$2\% (to
$34{,}373$); both remain non-functional. The operator matrices alone beat
random init ($35$k vs.\ $66$k) once the surrounding tensors are real, unlike
the fully projected model of \S\ref{sec:projection}, again locating the destruction
in the dense projection, not the residual. (The correction's gains on the 18
training layers are partly memorized; only the held-out layers carry genuine
predictions, and there the reduction is zero.)

\section{Training protocol details}
\label{app:protocol}

Every conversion arm within an experiment shares \emph{identical} data order,
optimizer (AdamW, $\beta=(0.9,0.95)$, weight decay $0.1$), cosine schedule
with 100-step warmup, gradient clipping at $1.0$, bf16 autocast, and token
budget (30M primary; 100M persistence checks); the learning rate equals the
target size's original pretraining rate. Initialization construction costs
are negligible against any training budget (selection: norms plus top-$k$,
seconds; moments pass for compensation over 1{,}000 frozen-corpus sequences:
$\sim$75\,s; closed-form solves: seconds; rescale: free) and are included in
the compute accounting. Every arm then trains on the same 30M tokens under
the same schedule, so all standings compare equal-budget endpoints differing
only in starting weights. The Pythia runs used a single NVIDIA GB10 (128\,GB
unified memory; $\sim$9.4k tok/s, $\sim$55\,min per 30M-token arm; the 1B
seed re-runs used cloud A100s), and the Qwen3 race used cloud NVIDIA A10G
GPUs (24\,GB), one arm per GPU, with the $32\times1024$ batch accumulated
over 16 micro-batches to hold the 151{,}936-way logits. Any 24\,GB+ CUDA GPU
reproduces every run.

\end{document}